\documentclass[10pt,twocolumn,letterpaper]{article}

\usepackage[table]{xcolor} 
\usepackage{cvpr}
\usepackage{times}
\usepackage{epsfig}
\usepackage{graphicx}
\usepackage{amsmath}
\usepackage{amssymb}

\usepackage[pagebackref=true,breaklinks=true,letterpaper=true,colorlinks,bookmarks=false]{hyperref}

\cvprfinalcopy 

\def\cvprPaperID{1534} 

\ifcvprfinal\fi

\newcommand{\PAR}[1]{\vskip4pt \noindent {\bf #1~}}  

\newcommand{\mytt}[1]{{\tt \footnotesize #1}}

\usepackage{ifthen}
\usepackage{csvsimple}
\usepackage{tabularx}
\usepackage{booktabs}
\usepackage{multirow}
\usepackage{siunitx}
\usepackage{pgfplots}
\usepackage{tikz}
\usepackage{caption}
\usepackage{subcaption}  
\usepackage[export]{adjustbox}

\makeatletter
\def\instring#1#2{TT\fi\begingroup
  \edef\x{\endgroup\noexpand\in@{#1}{#2}}\x\ifin@}
\makeatother

\makeatletter
\def\instring#1#2{TT\fi\begingroup
  \edef\x{\endgroup\noexpand\in@{#1}{#2}}\x\ifin@}
\makeatother

\newcommand{\fruc}[3]{\mathrel{#1\raisebox{2pt}{$\frac{#2}{#3}$}}\hspace*{-3pt}}

\definecolor{highlight_green} {RGB}{180,255,130}
\definecolor{better_blue}{HTML}{348ABD}
\definecolor{better_red}{HTML}{E24A33}
\definecolor{better_green}{HTML}{8EBA42}

\newcommand{\nb}[3]{\if\instring{b}{#2}\bfseries\else\fi\if\instring{i}{#2}\itshape{\bfseries}\else\fi\if\instring{h}{#2}\cellcolor{#3}\else\fi#1\if\instring{w}{#2}\phantom{0}\else\fi}
\newcommand{\np}[3]{\if\instring{p}{#2}\bfseries\else\fi\if\instring{i}{#2}\itshape{\bfseries}\else\fi\if\instring{h}{#2}\cellcolor{#3}\else\fi#1\if\instring{w}{#2}\phantom{0}\else\fi}

\pgfplotsset{
    discard if not/.style 2 args={
        x filter/.code={
            \edef\tempa{\thisrow{#1}}
            \edef\tempb{#2}
            \ifx\tempa\tempb
            \else
                
            \fi
        }
    }
}

\newcolumntype{C}{>{\centering\arraybackslash}X}

\newcommand{\norm}[1]{\left\lVert#1\right\rVert}  
\newcommand{\loss}[1]{\mathcal{L}_\textnormal{#1}}
\newcommand{\lossnnz}[1]{\mathcal{L}_{\textnormal{#1} \neq 0}}
\newcommand{\nnfn}{f_\theta}


\begin{document}

\title{In Defense of the Triplet Loss for Person Re-Identification}

\author{Alexander Hermans\thanks{Equal contribution. Ordering determined by a last minute coin flip.},\hspace*{2pt} Lucas Beyer\footnotemark[1]\hspace*{4pt} and Bastian Leibe\\
Visual Computing Institute\\
RWTH Aachen University\\
{\tt\small last@vision.rwth-aachen.org}
}

\maketitle

\begin{abstract}
    In the past few years, the field of computer vision has gone through a revolution fueled mainly by the advent of large datasets and the adoption of deep convolutional neural networks for end-to-end learning.
    The person re-identification subfield is no exception to this.
    Unfortunately, a prevailing belief in the community seems to be that the triplet loss is inferior to using surrogate losses (classification, verification) followed by a separate metric learning step.
    We show that, for models trained from scratch as well as pretrained ones, using a variant of the triplet loss to perform end-to-end deep metric learning outperforms most other published methods by a large margin.
\end{abstract}

\section{Introduction}
\begin{figure}[t]
        \includegraphics[width=\linewidth]{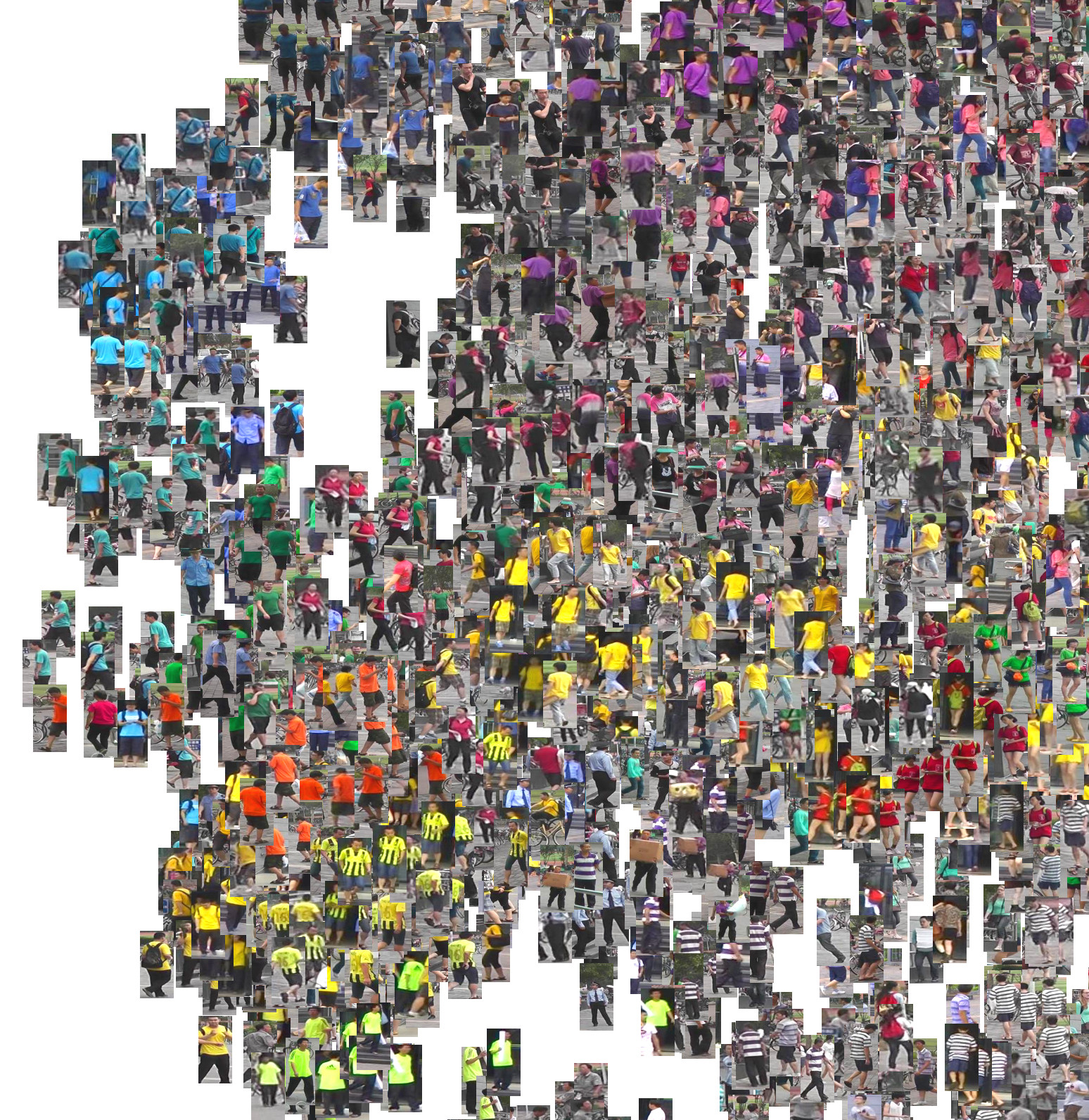}
        \caption{%
            A small crop of the Barnes-Hut t-SNE~\cite{Maaten14JMLR} of our learned embeddings for the Market-1501 test-set.
            The triplet loss learns semantically meaningful features.
        }
        \label{fig:tsne}
    \end{figure}

    In recent years, person re-identification (ReID) has attracted significant attention in the computer vision community.
    Especially with the rise of deep learning, many new approaches have been proposed to achieve this task~\cite{Wang16CVPR,Cheng16CVPR,Xiao16CVPR,Shi16ECCV,Varior16aECCV,Geng16Arxiv,ZhengZ16Arxiv,Barbosa17Arxiv,ZhangW17Arxiv,Li17CVPR,ZhengZ17ICCV,Sun17ICCV}\footnote{A nice overview of the field is given by a recent survey paper~\cite{ZhengL16Arxiv}.}.
    In many aspects person ReID is similar to image retrieval, where significant progress has been made and where deep learning has recently introduced a lot of changes.
    One prominent example in the recent literature is FaceNet~\cite{Schroff15CVPR}, a convolutional neural network (CNN) used to learn an embedding for faces.
    The key component of FaceNet is to use the triplet loss, as introduced by Weinberger and Saul~\cite{Weinberger09JMLR}, for training the CNN as an embedding function.
    The triplet loss optimizes the embedding space such that data points with the same identity are closer to each other than those with different identities.
    A visualization of such an embedding is shown in Figure~\ref{fig:tsne}.

    Several approaches for person ReID have already used some variant of the triplet loss to train their models~\cite{Khamis14ECCV,Ding15PR,Paisitkriangkrai15CVPR,Cheng16CVPR,Wang16CVPR,Shi16ECCV,Su16ECCV,Liu16MM,Chen16AAAI,Liu17TIP}, with moderate success.
    The recently most successful person ReID approaches argue that a classification loss, possibly combined with a verification loss, is superior for the task~\cite{Chen16AAAI,ZhengL16Arxiv,Geng16Arxiv,ZhengZ16Arxiv,Li17IJCAI}.
    Typically, these approaches train a deep CNN using one or multiple of these surrogate losses and subsequently use a part of the network as a feature extractor, combining it with a metric learning approach to generate final embeddings.
    Both of these losses have their problems, though.
    The classification loss necessitates a growing number of learnable parameters as the number of identities increases, most of which will be discarded after training.
    On the other hand, many of the networks trained with a verification loss have to be used in a cross-image representation mode, only answering the question ``How similar are these two images?''.
    This makes using them for any other task, such as clustering or retrieval, prohibitively expensive, as each probe has to go through the network paired up with every gallery image.

    In this paper we show that, contrary to current opinion, a plain CNN with a triplet loss can outperform current state-of-the-art approaches on the CUHK03~\cite{Li14CVPR}, Market-1501~\cite{ZhengL15ICCV} and MARS~\cite{ZhengL16ECCV} datasets.
    The triplet loss allows us to perform end-to-end learning between the input image and the desired embedding space.
    This means we directly optimize the network for the final task, which renders an additional metric learning step obsolete.
    Instead, we can simply compare persons by computing the Euclidean distance of their embeddings.

    A possible reason for the unpopularity of the triplet loss is that, when applied na\"ively, it will indeed often produce disappointing results.
    An essential part of learning using the triplet loss is the mining of hard triplets, as otherwise training will quickly stagnate.
    However, mining such hard triplets is time consuming and it is unclear what defines ``good'' hard triplets~\cite{Schroff15CVPR, Shi16ECCV}.
    Even worse, selecting too hard triplets too often makes the training unstable.
    We show how this problem can be alleviated, resulting in both faster training and better performance.
    We systematically analyze the design space of triplet losses, and evaluate which one works best for person ReID.
    While doing so, we place two previously proposed variants~\cite{Ding15PR,Song16CVPR} into this design space and discuss them in more detail in Section~\ref{sec:mle}.
    Specifically, we find that the best performing version has not been used before.
    Furthermore we also show that a margin-less formulation performs slightly better, while removing one hyper-parameter.

    Another clear trend seems to be the use of pretrained models such as GoogleNet~\cite{Szegedy15CVPR} or ResNet-50~\cite{He16CVPR}.
    Indeed, pretrained models often obtain great scores for person ReID~\cite{Geng16Arxiv,ZhengZ16Arxiv}, while ever fewer top-performing approaches use networks trained from scratch~\cite{Li14CVPR,Ahmed15CVPR,Cheng16CVPR,Xiao16CVPR,Shi16ECCV,Varior16aECCV,Barbosa17Arxiv}.
    Some authors even argue that training from scratch is bad~\cite{Geng16Arxiv}.
    However, using pretrained networks also leads to a design lock-in, and does not allow for the exploration of new deep learning advances or different architectures.
    We show that, when following best practices in deep learning, networks trained from scratch can perform competitively for person ReID.
    Furthermore, we do not rely on network components specifically tailored towards person ReID, but train a plain feed-forward CNN, unlike many other approaches that train from scratch~\cite{Li14CVPR,Ahmed15CVPR,Varior16aECCV,Xiao16CVPR,Subramaniam16NIPS,Li17CVPR,Zhao17CVPR}.
    Indeed, our networks using pretrained weights obtain the best results, but our far smaller architecture obtains respectable scores, providing a viable alternative for applications where person ReID needs to be performed on resource-constrained hardware, such as embedded devices.

    In summary our contribution is twofold:
    Firstly we introduce variants of the classic triplet loss which render mining of hard triplets unnecessary and we systematically evaluate these variants.
    And secondly, we show how, contrary to the prevailing opinion, using a triplet loss and no special layers, we achieve state-of-the-art results both with a pretrained CNN and with a model trained from scratch.
    This highlights that a well designed triplet loss has a significant impact on the result, on par with other architectural novelties, hopefully enabling other researchers to gain the full potential of the previously often dismissed triplet loss.
    This is an important result, highlighting that a well designed triplet loss has a significant impact on model performance---on par with other architectural novelties---hopefully enabling other researchers to gain the full potential of the previously often dismissed triplet loss.

%
%
%

\section{Learning Metric Embeddings, the Triplet Loss, and the Importance of Mining}\label{sec:mle}
    The goal of metric embedding learning is to learn a function $\nnfn(x) : \mathbb{R}^F \rightarrow \mathbb{R}^D$ which maps semantically similar points from the data manifold in $\mathbb{R}^F$ onto metrically close points in $\mathbb{R}^D$.
    Analogously, $\nnfn$ should map semantically different points in $\mathbb{R}^F$ onto metrically distant points in $\mathbb{R}^D$.
    The function $\nnfn$ is parametrized by $\theta$ and can be anything ranging from a linear transform~\cite{Weinberger09JMLR,Liao15CVPR,Zhang16CVPR,Paisitkriangkrai15CVPR} to complex non-linear mappings usually represented by deep neural networks~\cite{Ding15PR,Cheng16CVPR,Geng16Arxiv}.
    Let $D(x,y) : \mathbb{R}^D \times \mathbb{R}^D \rightarrow \mathbb{R}$ be a metric function measuring distances in the embedding space.
    For clarity we use the shortcut notation $D_{i,j} = D(\nnfn(x_i), \nnfn(x_j))$, where we omit the indirect dependence of $D_{i,j}$ on the parameters $\theta$.
    As is common practice, all loss-terms are divided by the number of summands in a batch; we omit this term in the following equations for conciseness.

    Weinberger and Saul~\cite{Weinberger09JMLR} explore this topic with the explicit goal of performing $k$-nearest neighbor classification in the learned embedding space and propose the ``Large Margin Nearest Neighbor loss'' for optimizing $\nnfn$:%
    \begin{equation}\label{eq:loss_lmnn}
        \loss{LMNN}(\theta) = (1 - \mu) \loss{pull}(\theta) + \mu \loss{push}(\theta),
    \end{equation}%
    which is comprised of a \textit{pull}-term, pulling data points $i$ towards their \textit{target neighbor} $T(i)$ from the same class,
    and a \textit{push}-term, pushing data points from a different class $k$ further away:%
    \begin{align}\label{eq:pullpush}
        \loss{pull}(\theta) &= \sum\limits_{i, j \in T(i)} D_{i,j}, \\
        \loss{push}(\theta) &= \sum\limits_{\substack{a,n \\ y_a \neq y_n}} \left[m + D_{a,T(a)} - D_{a,n}\right]_+.
    \end{align}%
    Because the motivation was nearest-neighbor classification, allowing disparate clusters of the same class was an explicit goal, achieved by choosing fixed target neighbors at the onset of training.
    Since this property is harmful for retrieval tasks such as face and person ReID, FaceNet~\cite{Schroff15CVPR} proposed a modification of $\loss{LMNN}(\theta)$ called the ``Triplet loss'':%
    \begin{equation}\label{eq:loss_trip}
        \loss{tri}(\theta) = \sum\limits_{\substack{a,p,n \\ y_a = y_p \neq y_n}} \left[m + D_{a,p} - D_{a,n}\right]_+.
    \end{equation}%
    This loss makes sure that, given an \textit{anchor} point $x_a$, the projection of a \textit{positive} point $x_p$ belonging to the same class (person) $y_a$ is closer to the anchor's projection than that of a \textit{negative} point belonging to another class $y_n$, by at least a margin $m$.
    If this loss is optimized over the whole dataset for long enough, eventually all possible pairs $(x_a, x_p)$ will be seen and be pulled together, making the pull-term redundant.
    The advantage of this formulation is that, while eventually all points of the same class will form a single cluster, they are not required to collapse to a single point; they merely need to be closer to each other than to any point from a different class.

    A major caveat of the triplet loss, though, is that as the dataset gets larger, the possible number of triplets grows cubically, rendering a long enough training impractical.
    To make matters worse, $\nnfn$ relatively quickly learns to correctly map most trivial triplets, rendering a large fraction of all triplets uninformative.
    Thus mining \textit{hard} triplets becomes crucial for learning.
    Intuitively, being told over and over again that people with differently colored clothes are different persons does not teach one anything, whereas seeing similarly-looking but different people (\textit{hard negatives}), or pictures of the same person in wildly different poses (\textit{hard positives}) dramatically helps understanding the concept of ``same person''.
    On the other hand, being shown \textit{only} the hardest triplets would select outliers in the data unproportionally often and make $\nnfn$ unable to learn ``normal'' associations, as will be shown in Table~\ref{tab:loss_baselines}.
    Examples of typical hard positives, hard negatives, and outliers are shown in the Supplementary Material.
    Hence it is common to only mine \textit{moderate} negatives~\cite{Schroff15CVPR} and/or \textit{moderate} positives~\cite{Shi16ECCV}.
    Regardless of which type of mining is being done, it is a separate step from training and adds considerable overhead, as it requires embedding a large fraction of the data with the most recent $\nnfn$ and computing all pairwise distances between those data points.

    In a classical implementation, once a certain set of $B$ triplets has been chosen, their images are stacked into a batch of size $3B$, for which the $3B$ embeddings are computed, which are in turn used to create $B$ terms contributing to the loss.
    Given the fact that there are up to $6B^2-4B$ possible combinations of these $3B$ images that are valid triplets, using only $B$ of them seems wasteful.
    With this realization, we propose an organizational modification to the classic way of using the triplet loss: the core idea is to form batches by randomly sampling $P$ classes (person identities), and then randomly sampling $K$ images of each class (person), thus resulting in a batch of $PK$ images.\footnote{\label{fn:pk}In all experiments we choose $B$, $P$, and $K$ in such a way that $3B$ is close to $PK$, \eg $3 \cdot 42 \approx 32 \cdot 4$.}
    Now, for each sample $a$ in the batch, we can select the hardest positive and the hardest negative samples \emph{within the batch} when forming the triplets for computing the loss, which we call \textit{Batch Hard}:%
    \begin{align}\label{eq:loss_bh}
        \loss{BH}(\theta; X) = \overbrace{\sum\limits_{i=1}^{P} \sum\limits_{a=1}^{K}}^{\textnormal{all anchors}}
            \Big[m & + \hspace*{-5pt} \overbrace{\max\limits_{p=1 \dots K} \hspace*{-5pt} D\left(\nnfn(x^i_a), \nnfn(x^i_p)\right)}^{\textnormal{hardest positive}} \\
                   & - \hspace*{-5pt} \underbrace{\min\limits_{\substack{j=1 \dots P \\ n=1 \dots K \\ j \neq i}} \hspace*{-5pt} D\left(\nnfn(x^i_a), \nnfn(x^j_n)\right)}_{\textnormal{hardest negative}} \Big]_+,\nonumber
    \end{align}%
    which is defined for a mini-batch $X$ and where a data point $x^i_j$ corresponds to the $j$-th image of the $i$-th person in the batch.

    This results in $PK$ terms contributing to the loss, a threefold\footnote{Because $PK \approx 3B$, see footnote~\ref{fn:pk}} increase over the traditional formulation.
    Additionally, the selected triplets can be considered \textit{moderate triplets}, since they are the hardest within a small subset of the data, which is exactly what is best for learning with the triplet loss.

    This new formulation of sampling a batch immediately suggests another alternative, that is to simply use all possible $PK(PK-K)(K-1)$ combinations of triplets, which corresponds to the strategy chosen in~\cite{Ding15PR} and which we call \textit{Batch All}:%
    \begin{align}\label{eq:loss_ba}
        \loss{BA}(\theta; X) &= \hspace*{-2pt}\overbrace{\sum\limits_{i=1}^{P} \sum\limits_{a=1}^{K}}^{\textnormal{all anchors}}
                                \overbrace{\sum\limits_{\substack{p=1 \\ p \neq a}}^{K}}^{\hspace*{3pt}\textnormal{all pos.}\hspace*{3pt}}
                                \overbrace{\sum\limits_{\substack{j=1 \\ j \neq i}}^{P} \sum\limits_{n=1}^{K}}^{\textnormal{all negatives}}
                                \Big[m + d^{i,a,p}_{j,a,n} \Big]_+,\\
        d^{i,a,p}_{j,a,n} &= D\left(\nnfn(x^i_a), \nnfn(x^i_p)\right) - D\left(\nnfn(x^i_a), \nnfn(x^j_n)\right).\nonumber
    \end{align}%
    At this point, it is important to note that both $\loss{BH}$ and $\loss{BA}$ still exactly correspond to the standard triplet loss in the limit of infinite training.
    Both the $\max$ and $\min$ functions are continuous and differentiable almost everywhere, meaning they can be used in a model trained by stochastic \mbox{(sub-)gradient} descent without concern.
    In fact, they are already widely available in popular deep-learning frameworks for the implementation of max-pooling and the ReLU~\cite{Glorot11AISTATS} non-linearity.

    Most similar to our \textit{batch hard} and \textit{batch all} losses is the \textit{Lifted Embedding} loss~\cite{Song16CVPR}, which fills the batch with triplets but considers all but the anchor-positive pair as negatives:%
    \begin{equation}\label{eq:loss_lift}
        \loss{L}(\theta; X) = \hspace*{-10pt} \sum\limits_{(a,p) \in X} \hspace*{-6pt}
            \Big[D_{a,p} + \log\hspace*{-10pt}\sum\limits_{\substack{n \in X \\ n \neq a, n \neq p}}\hspace*{-10pt}\left(e^{m - D_{a,n}} + e^{m - D_{p,n}} \right)\Big]_+.\nonumber
    \end{equation}%
    While~\cite{Song16CVPR} motivates a ``hard''-margin loss similar to $\loss{BH}$ and $\loss{BA}$, they end up optimizing the smooth bound of it given in the above equation.
    Additionally, traditional $3B$ batches are considered, thus using all possible negatives, but only one positive pair per triplet.
    This leads us to propose a generalization of the Lifted Embedding loss based on $PK$ batches which considers all anchor-positive pairs as follows:%
    \begin{align}\label{eq:loss_lift_npos}
        \loss{LG}(\theta; X) = \overbrace{\sum\limits_{i=1}^{P} \sum\limits_{a=1}^{K}}^{\textnormal{all anchors}}
            \Big[\log \overbrace{\sum\limits_{\substack{p=1 \\ p \neq a}}^{K} e^{D\left(\nnfn(x^i_a), \nnfn(x^i_p)\right)}}^{\textnormal{all positives}}
            \\
            + \log\underbrace{\sum\limits_{\substack{j=1 \\ j \neq i}}^{P} \sum\limits_{n=1}^{K} e^{m - D\left(\nnfn(x^i_a), \nnfn(x^j_n)\right)}}_{\textnormal{all negatives}}\Big]_+.\nonumber
    \end{align}%

    \PAR{Distance Measure.}
    Throughout this section, we have referred to $D(a,b)$ as the distance function between $a$ and $b$ in the embedding space.
    In most related works, the squared Euclidean distance $D\left(\nnfn(x_i), \nnfn(x_j)\right) = \norm{\nnfn(x_i) - \nnfn(x_j)}_2^2$ is used as metric, although nothing in the above loss definitions precludes using any other \mbox{(sub-)differentiable} distance measure.
    While we do not have a side-by-side comparison, we noticed during initial experiments that using the squared Euclidean distance made the optimization more prone to collapsing, whereas using the actual (non-squared) Euclidean distance was more stable.
    We hence used the Euclidean distance throughout all our experiments presented in this paper.
    In addition, squaring the Euclidean distance makes the margin parameter less interpretable, as it does not represent an absolute distance anymore.

    Note that when forcing the embedding's norm to one, using the squared Euclidean distance corresponds to using the cosine-similarity, up to a constant factor of two.
    We did not use a normalizing layer in any of our final experiments.
    For one, it does not dramatically regularize the network by reducing the available embedding space: the space spanned by all $D$-dimensional vector of fixed norm is still a $D-1$-dimensional volume.
    Worse, an output-normalization layer can actually hide problems in the training, such as slowly collapsing or exploding embeddings.

    \PAR{Soft-margin.}
    The role of the hinge function $\left[m + \bullet\right]_+$ is to avoid correcting ``already correct'' triplets.
    But in person ReID, it can be beneficial to pull together samples from the same class as much as possible~\cite{Zhang16CVPR,Cheng16CVPR}, especially when working on tracklets such as in MARS~\cite{ZhengL16ECCV}.
    For this purpose, it is possible to replace the hinge function by a smooth approximation using the softplus function: $\ln(1 + \exp(\bullet))$,
%
%
    for which numerically stable implementations are commonly available as \mytt{log1p}.
    The softplus function has similar behavior to the hinge, but it decays exponentially instead of having a hard cut-off, we hence refer to it as the soft-margin formulation.

    \PAR{Summary.}
    In summary, the novel contributions proposed in this paper are the \textit{batch hard} loss and its soft margin version.
    In the following section we evaluate them experimentally and show that, for ReID, they achieve superior performance compared to both the traditional triplet loss and the previously published variants of it~\cite{Ding15PR,Song16CVPR}.

\section{Experiments}
    Our experimental evaluation is split up into three main parts.
    The first section evaluates different variations of the triplet loss, including some hyper-parameters, and identifies the setting that works best for person ReID.
    This evaluation is performed on a train/validation split we create based on the MARS training set.
    The second section shows the performance we can attain based on the selected variant of the triplet loss.
    We show state-of-the-art results on the CUHK03, Market-1501 and MARS test sets, based on a pretrained network and a network trained from scratch.
    Finally, the third section discusses advantages of training models from scratch with respect to real-world use cases.

    \subsection{Datasets}
        We focus on the Market-1501~\cite{ZhengL15ICCV} and MARS~\cite{ZhengL16ECCV} datasets, the two largest person ReID datasets currently available.
        The Market-1501 dataset contains bounding boxes from a person detector which have been selected based on their intersection-over-union overlap with manually annotated bounding boxes.
        It contains $\num{32668}$ images of $\num{1501}$ persons, split into train/test sets of $\num{12936}$/$\num{19732}$ images as defined by~\cite{ZhengL15ICCV}.
        The dataset uses both single- and multi-query evaluation, we report numbers for both.

        The MARS dataset originates from the same raw data as the Market-1501 dataset; however, a significant difference is that the MARS dataset does not have any manually annotated bounding boxes, reducing the annotation overhead.
        MARS consist of ``tracklets'' which have been grouped into person IDs manually.
        It contains $\num{1191003}$ images split into train/test sets of $\num{509914}$/$\num{681089}$ images, as defined by~\cite{ZhengL16ECCV}.
        Here, person ReID is no longer performed on a frame-to-frame level, but instead on a tracklet-to-tracklet level, where feature embeddings are pooled across a tracklet, thus it is inherently a multi-query setup.

        We use the standard evaluation metrics for both datasets, namely the mean average precision score (mAP) and the cumulative matching curve (CMC) at rank-1 and rank-5.
        To compute these scores we use the evaluation code provided by~\cite{Zhong17CVPR}.

        Additionally, we show results on the CUHK03~\cite{Li14CVPR} dataset for our pretrained network, using the single shot setup and average over the provided 20 train/test splits.

        \newcommand{\addlosstableline}[2]{\csvreader[ column count=21, late after line=\\, filter=\equal{\description}{#1}]
                                                    {data/loss_baselines.csv}
                                                    {1=\description,  2=\zeroOneMap,   3=\zeroOneMapBf,
                                                                      4=\zeroOneRank,  5=\zeroOneRankBf,
                                                                      6=\zeroTwoMap,   7=\zeroTwoMapBf,
                                                                      8=\zeroTwoRank,  9=\zeroTwoRankBf,
                                                                     10=\zeroFiveMap,  11=\zeroFiveMapBf,
                                                                     12=\zeroFiveRank, 13=\zeroFiveRankBf,
                                                                     14=\oneZeroMap,   15=\oneZeroMapBf,
                                                                     16=\oneZeroRank,  17=\oneZeroRankBf,
                                                                     18=\softMap,      19=\softMapBf,
                                                                     20=\softRank,     21=\softRankBf}
                                                    {#2 &  \nb{\zeroOneMap}{\zeroOneMapBf}{highlight_green}   && \nb{\zeroOneRank}{\zeroOneRankBf}{highlight_green}
                                                        && \nb{\zeroTwoMap}{\zeroTwoMapBf}{highlight_green}   && \nb{\zeroTwoRank}{\zeroTwoRankBf}{highlight_green}
                                                        && \nb{\zeroFiveMap}{\zeroFiveMapBf}{highlight_green} && \nb{\zeroFiveRank}{\zeroFiveRankBf}{highlight_green}
                                                        && \nb{\oneZeroMap}{\oneZeroMapBf}{highlight_green}   && \nb{\oneZeroRank}{\oneZeroRankBf}{highlight_green}
                                                        && \nb{\softMap}{\softMapBf}{highlight_green}         && \nb{\softRank}{\softRankBf}{highlight_green}
                                                    }
                                         }

        \begin{table*}[t]
            \setlength{\tabcolsep}{1pt}
            \setlength{\extrarowheight}{5pt}
            \renewcommand{\arraystretch}{0.75}
            \centering
            \begin{tabularx}{\textwidth}{p{3.5cm}Cp{2pt}Cp{13pt}Cp{2pt}Cp{13pt}Cp{2pt}Cp{13pt}Cp{2pt}Cp{13pt}Cp{2pt}C}
                \toprule[1pt]
                 & \multicolumn{3}{c}{margin 0.1} && \multicolumn{3}{c}{margin 0.2} && \multicolumn{3}{c}{margin 0.5} && \multicolumn{3}{c}{margin 1.0} && \multicolumn{3}{c}{soft margin} \\
                \cmidrule[0.5pt]{2-4} \cmidrule[0.5pt]{6-8} \cmidrule[0.5pt]{10-12} \cmidrule[0.5pt]{14-16} \cmidrule[0.5pt]{18-20}
                 & mAP && rank-1 && mAP && rank-1 && mAP && rank-1 && mAP && rank-1 && mAP && rank-1 \\
                \midrule[0.5pt]
                \addlosstableline{triplet}{Triplet ($\loss{tri}$)}
                \addlosstableline{triplet + HN}{Triplet ($\loss{tri}$) + OHM}
                \addlosstableline{all pairs}{Batch hard ($\loss{BH}$)}
                \addlosstableline{all pairs + nnz}{Batch hard ($\lossnnz{BH}$)}
                \addlosstableline{for real}{Batch all ($\loss{BA}$)}
                \addlosstableline{for real + nnz}{Batch all ($\lossnnz{BA}$)}
                \addlosstableline{lifted 3pos}{Lifted $3$-pos. ($\loss{LG}$)}
                \addlosstableline{lifted 1pos}{Lifted $1$-pos. ($\loss{L}$)~\cite{Song16CVPR}}
                \bottomrule[1pt]
            \end{tabularx}
            \caption{\textit{LuNet} scores on our MARS validation split.
            The best performing loss at a given margin is bold, the best margin for a given loss is italic, and the overall best combination is highlighted in green.
            A * denotes runs trapped in a bad local optimum.}
            \label{tab:loss_baselines}
            \vspace*{-11pt}
        \end{table*}

    \subsection{Training}
        Unless specifically noted otherwise, we use the same training procedure across all experiments and on all datasets.
        We performed all our experiments using the Theano~\cite{Bastien12NIPSW} framework, code is available at \mytt{redacted}.

        We use the Adam optimizer~\cite{Kingma15ICLR} with the default hyper-parameter values ($\epsilon=10^{-3}$, $\beta_1=0.9$, $\beta_2=0.999$) for most experiments.
        During initial experiments on our own MARS validation split (see Sec.~\ref{sec:loss_baselines}), we ran multiple experiments for a very long time and monitored the loss and mAP curves.
        With this information, we decided to fix the following exponentially decaying training schedule, which does not disadvantage any setup, for all experiments presented in this paper:%
        \begin{equation}
            \epsilon(t) = \begin{cases}
                \epsilon_0                                          & \text{if } t \leq t_0 \\
                \epsilon_0 \num{0.001}^{\frac{t-t_0}{t_1-t_0}}      & \text{if } t_0 \leq t \leq t_1
            \end{cases}
        \end{equation}%
        with $\epsilon_0 = 10^{-3}$, $t_0 = \num{15000}$, and $t_1 = \num{25000}$, stopping training when reaching $t_1$.
        We also set $\beta_1 = 0.5$ when entering the decay schedule at $t_0$, as is common practice~\cite{Ba16Arxiv}.

        In the Supplementary Material, we provide a detailed discussion of various interesting effects we regularly observed during training, providing hands-on guidance for other researchers.

    \subsection{Network Architectures}\label{sec:nets}
        For our main results we use two different architectures, one based on a pretrained network and one which we train from scratch.

        \PAR{Pretrained.}
        We use the ResNet-50 architecture and the weights provided by He~\etal~\cite{He16CVPR}.
        We discard the last layer and add two fully connected layers for our task.
        The first has 1024 units, followed by batch normalization~\cite{Ioffe15ICML} and ReLU~\cite{Glorot11AISTATS}, the second goes down to 128 units, our final embedding dimension.
        Trained with our \textit{batch hard} triplet loss, we call this model \textit{TriNet}.
        Due to the size of this network (\SI{25.74}{M} parameters), we had to limit our batch size to 72, containing $P=18$ persons with $K=4$ images each.
        For these pretrained experiments, $\epsilon_0 = 10^{-3}$ proved to be too high, causing the models to diverge within few iterations.
        We thus reduced $\epsilon_0$ to $3 \cdot 10^{-4}$ which worked fine on all datasets.

        \PAR{Trained from Scratch.}
        To show that training from scratch does not necessarily result in poor performance, we also designed a network called \textit{LuNet} which we train from scratch.
        \textit{LuNet} follows the style of ResNet-v2, but uses leaky ReLU nonlinearities, multiple $3 \times 3$ max-poolings with stride 2 instead of strided convolutions, and omits the final average-pooling of feature-maps in favor of a channel-reducing final res-block.
        An in-depth description of the architecture is given in the Supplementary Material.
        As the network is much more lightweight (\SI{5.00}{M} parameters) than its pretrained sibling, we sample batches of size 128, containing $P=32$ persons with $K=4$ images each.

    \subsection{Triplet Loss}\label{sec:loss_baselines}
        Our initial experiments test the different variants of triplet training that we discussed in Sec.~\ref{sec:mle}.
        In order not to perform model-selection on the test set, we randomly sample a validation set of 150 persons from the MARS training set, leaving the remaining 475 persons for training.
        In order to make this exploration tractable, we run all of these experiments using the smaller \textit{LuNet} trained from scratch on images downscaled by a factor of two.
        Since our goal here is to explore triplet loss formulations, as opposed to reaching top performance, we do not perform any data augmentation in these experiments.

        Table~\ref{tab:loss_baselines} shows the resulting mAP and rank-1 scores for the different formulations at multiple margin values, and with a soft-margin where applicable.
        Consistent with results reported in several recent papers~\cite{Geng16Arxiv,Ding15PR,Chen16AAAI}, the vanilla triplet loss with randomly sampled triplets performs poorly.
        When performing simple offline hard-mining (OHM), the scores sometimes increase dramatically, but the training also fails to learn useful embeddings for multiple margin values.
        This problem is well-known~\cite{Schroff15CVPR,Shi16ECCV} and has been discussed in Sec.~\ref{sec:mle}.
        While the idea of learning embeddings using triplets is theoretically pleasing, this practical finnickyness, coupled with the considerable increase in training time due to non-parallelizable offline mining (from \SI{7}{h} to \SI{20}{h} in our experiments), makes learning with vanilla triplets rather unattractive.

        Considering the long training times, it is nice to see that all proposed triplet re-formulations perform similarly to or better than the best OHM run.
        The key observation is that the (semi) hard-mining happens within the batch and thus comes at almost no additional runtime cost.

        Perhaps surprisingly, the \textit{batch hard} variant (Eq.~\ref{eq:loss_bh}) consistently outperforms the \textit{batch all} variant (Eq.~\ref{eq:loss_ba}) previously used by several authors~\cite{Ding15PR,Wang16CVPR}.
        We suspect this is due to the fact that in the latter, many of the possible triplets in a batch are zero, essentially ``washing out'' the few useful contributing terms during averaging.
        To test this hypothesis, we also ran experiments where we only average the non-zero loss terms (marked by $\neq 0$ in Table~\ref{tab:loss_baselines}); this performs much better in the \textit{batch all} case.
        Another interpretation of this modification is that it dynamically increases the weight of triplets which remain active as they get fewer.

        The lifted triplet loss $\loss{L}$ as introduced by~\cite{Song16CVPR} performs competitively, but is slightly worse than most other formulations.
        As can be seen in the table, our generalization to multiple positives (Eq.~\ref{eq:loss_lift_npos}), which makes it more similar to the \textit{batch all} variant of the triplet, improves upon it overall.

        The best score was obtained by the soft-margin variation of the \textit{batch hard} loss.
        We use this loss in all our further triplet experiments.
        To clarify, here we merely seek the best triplet loss variation for person ReID, but do not claim that this variant works best across all fields.
        For other tasks such as image retrieval or clustering, additional experiments will have to be performed.

        \newcommand{\addmarskettableline}[3]{\csvreader[ column count=21, late after line=\\, filter=\equal{\description}{#1}]
                                                    {data/marsket_data.csv}
                                                    {1=\description,  2=\singQueryMAP,       3=\singQueryMAPBf,
                                                                      4=\singQueryRankOne,   5=\singQueryRankOneBf,
                                                                      6=\singQueryRankFive,  7=\singQueryRankFiveBf,
                                                                      8=\multQueryMAP,       9=\multQueryMAPBf,
                                                                     10=\multQueryRankOne,   11=\multQueryRankOneBf,
                                                                     12=\multQueryRankFive,  13=\multQueryRankFiveBf,
                                                                     14=\marsMAP,            15=\marsMAPBf,
                                                                     16=\marsRankOne,        17=\marsRankOneBf,
                                                                     18=\marsRankFive,       19=\marsRankFiveBf}
                                                    {#2 & #3
                                                     &  \nb{\singQueryMAP}{\singQueryMAPBf}{}
                                                     && \nb{\singQueryRankOne}{\singQueryRankOneBf}{}
                                                     && \nb{\singQueryRankFive}{\singQueryRankFiveBf}{}
                                                     && \nb{\multQueryMAP}{\multQueryMAPBf}{}
                                                     && \nb{\multQueryRankOne}{\multQueryRankOneBf}{}
                                                     && \nb{\multQueryRankFive}{\multQueryRankFiveBf}{}
                                                     && \nb{\marsMAP}{\marsMAPBf}{}
                                                     && \nb{\marsRankOne}{\marsRankOneBf}{}
                                                     && \nb{\marsRankFive}{\marsRankFiveBf}{}
                                                    }
                                            }

        \begin{table*}[t]
            \setlength{\tabcolsep}{1pt}
            \setlength{\extrarowheight}{5pt}
            \renewcommand{\arraystretch}{0.75}
            \centering
            \begin{tabularx}{\textwidth}{p{4.2cm}cCp{1pt}Cp{1pt}Cp{8pt}Cp{1pt}Cp{1pt}Cp{18pt}Cp{1pt}Cp{1pt}C}
                \toprule[1pt]
                & \multirow{2}{25pt}{\hspace*{8pt}\rotatebox{90}{\hspace*{-10pt}Type}} & \multicolumn{5}{c}{Market-1501 SQ} && \multicolumn{5}{c}{Market-1501 MQ} && \multicolumn{5}{c}{MARS}\\
                \cmidrule[0.5pt]{3-7} \cmidrule[0.5pt]{9-13} \cmidrule[0.5pt]{15-19}
                 && mAP && rank-1 && rank-5 && mAP && rank-1 && rank-5 && mAP && rank-1 && rank-5\\
                \midrule[0.5pt]
                \addmarskettableline{LaNet+flipcrop}{TriNet}{E}
                \addmarskettableline{LuNet+flipcrop}{LuNet}{E}
                \addmarskettableline{classification+avg+flipcrop+xqda}{IDE (R) + ML ours }{I}
                \arrayrulecolor{lightgray}\midrule[0.25pt]\arrayrulecolor{black}
                \addmarskettableline{nullspace}{LOMO + Null Space~\cite{Zhang16CVPR}}{E}
                \addmarskettableline{gated siamese}{Gated siamese CNN~\cite{Varior16aECCV}}{V}
                \addmarskettableline{CAN}{CAN~\cite{Liu17TIP}}{E}
                \addmarskettableline{multi-loss}{JLML~\cite{Li17IJCAI}}{I}
                \addmarskettableline{discriminLZ}{ResNet 50 (I+V)$^\dagger$~\cite{ZhengZ16Arxiv}}{I+V}
                \addmarskettableline{DTL Trip}{DTL$^\dagger$~\cite{Geng16Arxiv}}{E}
                \addmarskettableline{DTL SID+PV}{DTL$^\dagger$~\cite{Geng16Arxiv}}{I+V}
                \addmarskettableline{attributes}{APR (R, 751)$^\dagger$~\cite{Lin17Arxiv}}{I}
                \addmarskettableline{latent parts}{Latent Parts (Fusion)~\cite{Li17CVPR}}{I}
                \addmarskettableline{reranking baseline}{IDE (R) + ML~\cite{Zhong17CVPR}}{I}
                \addmarskettableline{spatial temporal RNN}{spatial temporal RNN~\cite{Zhou17CVPR}}{E}
                \addmarskettableline{compact appearance}{CNN + Video$^\dagger$~\cite{ZhangW17Arxiv}}{I}
                \arrayrulecolor{lightgray}\midrule[0.25pt]\arrayrulecolor{black}
                \addmarskettableline{LaNet+flipcrop+rerank}{TriNet (Re-ranked)}{E}
                \addmarskettableline{LuNet+flipcrop+rerank}{LuNet (Re-ranked)}{E}
                \addmarskettableline{classification+avg+flipcrop+xqda+rerank}{IDE (R) + ML ours (Re-ra.)}{I}
                \addmarskettableline{reranking}{IDE (R) + ML (Re-ra.)~\cite{Zhong17CVPR}}{I}
                \bottomrule[1pt]
            \end{tabularx}
            \caption{%
                Scores on both the Market-1501 and MARS datasets.
                The top and middle contain our scores and those of the current state-of-the-art respectively.
                The bottom contains several methods with re-ranking~\cite{Zhong17CVPR}.
                The different types represent the optimization criteria, where I stands for identification, V for verification and E for embedding.
                All our scores include test-time augmentation.
                The best scores are bold.
                $^\dagger$: Concurrent work only published on arXiv.
            }
            \label{tab:marsket_results}
            \vspace*{-11pt}
        \end{table*}

    \subsection{Performance Evaluation}
        Here, we present the main experiments of this paper.
        We perform all following experiments using the \textit{batch hard} variant $\loss{BH}$ of the triplet loss and the soft margin, since this setup performed best during the exploratory phase.

        \PAR{Batch Generation and Augmentation.}
        Since our \textit{batch hard} triplet loss requires slightly different mini-batches, we sample random $PK$-style batches by first randomly sampling $P$ person identities uniformly without replacement.
        For each person, we then sample $K$ images, without replacement whenever possible, otherwise replicating images as necessary.

        We follow common practice by using random crops and random horizontal flips during training~\cite{Krizhevsky12NIPS,Ahmed15CVPR}.
        Specifically, we resize all images of size $H \times W$ to $\scriptsize\fruc{1}{1}{8}(H \times W)$, of which we take random crops of size $H \times W$, keeping their aspect ratio intact.
        For all pretrained networks we set $H=256, W=128$ on Market-1501 and MARS and $H=256, W=96$ on CUHK03, whereas for the networks trained from scratch we set $H=128, W=64$.

        We apply test-time augmentation in all our experiments.
        Following~\cite{Krizhevsky12NIPS}, we deterministically average over the embeddings from five crops and their flips.
        This typically gives an improvement of $3\%$ in the mAP score; a more detailed analysis can be found in the Supplementary Material.

        \PAR{Combination of Embeddings.}
        For test-time augmentation, multi-query evaluation, and tracklet-based evaluation, the embeddings of multiple images need to be combined.

        While the learned clusters have no reason to be Gaussian, their convex hull is trained to only contain positive points.
        Thus, a convex combination of multiple embeddings cannot get closer to a negative embedding than any of the original ones, which is not the case for a non-convex combination such as max-pooling.
        For this reason, we suggest combining triplet-based embeddings by using their mean.
        For example, combining tracklet-embeddings using max-pooling led to an $11.4\%$ point decrease in mAP on MARS.

        \PAR{Comparison to State-of-the-Art.}
        Tables~\ref{tab:marsket_results} and~\ref{tab:cuhk_results} compare our results to a set of related, top performing approaches on Market-1501 and MARS, and CUHK03, respectively.
        We do not include approaches which are orthogonal to ours and could be integrated in a straightforward manner, such as various re-ranking schemes, data augmentation, and regularization~\cite{Bai17CVPR,Yu17BMVC,ZhengZ17ICCV,Chen17ICCVW,ZhangY17Arxiv,ZhengZ17Arxiv,Zhong17Arxiv}.
        These are included in more exhaustive tables in the Supplementary Material.
        The different approaches are categorized into three major types: Identification models (I) that are trained to classify person IDs, Verification models (V) that learn whether an image pair represents the same person, and methods such as ours that directly learn an Embedding (E).

        We present results for both our pretrained network (TriNet) and the network we trained from scratch (LuNet).
        As can clearly be seen, TriNet outperforms all current methods.
        Especially striking is the jump from $41.5\%$ mAP, obtained by another ResNet-50 model trained with triplets (DTL~\cite{Geng16Arxiv}), to our $69.14\%$ mAP score in Table~\ref{tab:marsket_results}.
        Since Geng~\etal~\cite{Geng16Arxiv} do not discus all details of their training procedure when using the triplet loss, we could only speculate about the reasons for the large performance gap.

        Our LuNet model, which we train from scratch, also performs very competitively, matching or outperforming most other baselines.
        While it does not quite reach the performance of our pretrained model, our results clearly show that with proper training, the flexibility of training models from scratch (see Sec.~\ref{sec:pretrain_scratch}) should not be discarded.

        To show that the actual performance boost is indeed gained by the triplet loss and not by other design choices, we train a ResNet-50 model with a classification loss.
        This model is very similar to the one used in~\cite{Zhong17CVPR} and we thus refer to it as ``IDE (R) ours'', for which we also apply a metric learning step (XQDA~\cite{Liao15CVPR}).
        Unfortunately, especially difficult images caused frequent spikes in the loss, which ended up harming the optimization using Adam.
        After unsuccessfully trying lower learning rates and clipping extreme loss values, we resorted to Adadelta~\cite{Zeiler12Arxiv}, another competitive optimization algorithm which did not exhibit these problems.
        While we combine embeddings through average pooling for our triplet based models, we found max-pooling and normalization to work better for the classification baseline, consistent with results reported in~\cite{ZhengL16ECCV}.
        As Table~\ref{tab:marsket_results} shows, the performance of the resulting model ``IDE (R) ours'' is still on-par with similar models in the literature.
        However, the large gap between the identification-based model and our TriNet clearly demonstrates the advantages of using a triplet loss.

        In line with the general trend in the vision community, all deep learning methods outperform shallow methods using hand-crafted features.
        While Table~\ref{tab:marsket_results} only shows~\cite{Zhang16CVPR} as a non-deep learning method, to the best of our knowledge all others perform worse.


        \newcommand{\addcuhktableline}[3]{\csvreader[ column count=9, late after line=\\, filter=\equal{\description}{#1}]
                                                    {data/cuhk03.csv}
                                                    {1=\description, 2=\labeledRankOne,   3=\labeledRankOneBf,
                                                                     4=\labeledRankFive,  5=\labeledRankFiveBf,
                                                                     6=\detectedRankOne,  7=\detectedRankOneBf,
                                                                     8=\detectedRankFive, 9=\detectedRankFiveBf}
                                                    {#2 & #3
                                                     &  \nb{\labeledRankOne}{\labeledRankOneBf}{}
                                                     && \nb{\labeledRankFive}{\labeledRankFiveBf}{}
                                                     && \nb{\detectedRankOne}{\detectedRankOneBf}{}
                                                     && \nb{\detectedRankFive}{\detectedRankFiveBf}{}
                                                    }
                                            }

        \begin{table}
            \small
            \setlength{\tabcolsep}{1pt}
            \setlength{\extrarowheight}{5pt}
            \renewcommand{\arraystretch}{0.75}
            \centering
            \begin{tabularx}{\linewidth}{p{3.7cm}cCp{1pt}Cp{8pt}Cp{1pt}C}
                \toprule[1pt]
                & \multirow{2}{25pt}{\hspace*{8pt}\rotatebox{90}{\hspace*{-10pt}Type}} & \multicolumn{3}{c}{Labeled} && \multicolumn{3}{c}{Detected}\\
                \cmidrule[0.5pt]{3-5} \cmidrule[0.5pt]{7-9}
                 && r-1 && r-5 && r-1 && r-5\\
                \midrule[0.5pt]
                \addcuhktableline{LaNet+flipcrop}{TriNet}{E}
                \arrayrulecolor{lightgray}\midrule[0.25pt]\arrayrulecolor{black}
                \addcuhktableline{gated siamese}{Gated siamese CNN~\cite{Varior16aECCV}}{V}
                \addcuhktableline{nullspace}{LOMO + Null Space~\cite{Zhang16CVPR}}{E}
                \addcuhktableline{CAN}{CAN~\cite{Liu17TIP}}{E}
                \addcuhktableline{latent parts}{Latent Parts (Fusion)~\cite{Li17CVPR}}{I}
                \addcuhktableline{spindle net}{Spindle Net*~\cite{Zhao17CVPR}}{I}
                \addcuhktableline{multi-loss}{JLML~\cite{Li17IJCAI}}{I}
                \addcuhktableline{DTL SID+PV}{DTL$^\dagger$~\cite{Geng16Arxiv}}{I+V}
                \addcuhktableline{discriminLZ}{ResNet 50 (I+V)$^\dagger$~\cite{ZhengZ16Arxiv}}{I+V}
                \bottomrule[1pt]
            \end{tabularx}
            \caption{%
                Scores on CUHK03 for TriNet and a set of recent top performing methods.
                The best scores are highlighted in bold.
                $^\dagger$: Concurrent work only published on arXiv.
                *: The method was trained on several additional datasets.
            }
            \label{tab:cuhk_results}
            \vspace*{-11pt}
        \end{table}

        We also evaluated how our models fare when combined with a recent re-ranking approach by Zhong~\etal~\cite{Zhong17CVPR}.
        This approach can be applied on top of any ranking methods and uses information from nearest neighbors in the gallery to improve the ranking result.
        As Table~\ref{tab:marsket_results} shows, our approaches go well with this method and show similar improvements to those obtained by Zhong~\etal~\cite{Zhong17CVPR}.

        Finally, we evaluate our models on Market-1501 with the provided $500k$ additional distractor images.
        The full experiment is described in the Supplementary Material.
        Even with these additional distractors, our triplet-based model outperforms a classification one by $8.4\%$ mAP.

        All of these results show that triplet loss embeddings are indeed a valuable tool for person ReID and we expect them to significantly change the way how research will progress in this field.

        \newcommand{\addimagesizedata}[1]{\csvreader[ column count=21, after line=, filter=\equal{\description}{#1}]
            {data/marsket_data.csv}
            {1=\description,  2=\singQueryMAP,
                              4=\singQueryRankOne,
                              6=\singQueryRankFive}
            {\nb{\singQueryMAP}{}{}
                && \nb{\singQueryRankOne}{}{}
                && \nb{\singQueryRankFive}{}{}
            }
        }

        \begin{table}[t]
            \footnotesize
            \setlength{\tabcolsep}{1pt}
            \setlength{\extrarowheight}{5pt}
            \renewcommand{\arraystretch}{0.75}
            \centering
            \begin{tabularx}{\linewidth}{p{1.6cm} Cp{2pt}Cp{2pt}C p{13pt} Cp{2pt}Cp{2pt}C}
                \toprule[1pt]
                 & \multicolumn{5}{c}{TriNet} && \multicolumn{5}{c}{LuNet}\\
                \cmidrule[0.5pt]{2-6} \cmidrule[0.5pt]{8-12}
                 & mAP && rank-1 && rank-5 && mAP && rank-1 && rank-5\\
                \midrule[0.5pt]
                $256 \times 128$ & \addimagesizedata{LaNet+flipcrop} && - && - && -\\
                $128 \times 64$ & \addimagesizedata{LaNet_half+flipcrop} && \addimagesizedata{LuNet+flipcrop}\\
                $64 \times 32$ & \addimagesizedata{LaNet_quarter+flipcrop} && \addimagesizedata{LuNet_half+flipcrop}\\
                \bottomrule[1pt]
            \end{tabularx}
            \caption{The effect of input size on mAP and CMC scores.}
            \label{tab:image_size}
            \vspace*{-11pt}
        \end{table}

    \subsection{To Pretrain or not to Pretrain?}\label{sec:pretrain_scratch}
        As mentioned before, many methods for person ReID rely on pretrained networks, following a general trend in the computer vision community.
        Indeed, these models lead to impressive results, as we also confirmed in this paper with our TriNet model.
        However, pretrained networks reduce the flexibility to try out new advances in deep learning or to make task-specific changes in a network.
        Our LuNet model clearly suggests that it is also possible to train models from scratch and obtain competitive scores.

        In particular, an interesting direction for ReID could be the usage of additional input channels such as depth information, readily available from cheap consumer hardware.
        However it is unclear how to best integrate such input data into a pretrained network in a proper way.

        Furthermore, the typical pretrained networks are designed with accuracy in mind and do not focus on the memory footprint or the runtime of a method.
        Both are important factors for real-world robotic scenarios, where typically power consumption is a constraint and only less powerful hardware can be considered~\cite{Hawes16RAM,Triebel16FSR}.
        When designing a network from scratch, one can directly take this into consideration and create networks with a smaller memory footprint and faster evaluation times.

        In principle, our pretrained model can easily be sped up by using half or quarter size input images, since the global average pooling in the ResNet will still produce an output vector of the same shape.
        This, however, goes hand in hand with the question of how to best adapt a pretrained network to a new task with different image sizes.
        The typical way of leveraging pretrained networks is to simply stretch images to the fixed expected input size used to train the network, typically $224 \times 224$ pixels.
        We used $256 \times 128$ instead in order to preserve the aspect ratio of the original image.
        However, for the Market-1501 dataset, this meant we had to upscale the images, while if we do not confine ourselves to pretrained networks we can simply adjust our architecture to the dataset size, as we did in the LuNet architecture.
        However, we hypothesize that a pretrained network has an ``intrinsic scale,'' for which the learned filters work properly and thus simply using smaller input images will result in suboptimal performance.
        To show this, we retrain our TriNet with $128 \times 64$ and $64 \times 32$ images.
        As Table~\ref{tab:image_size} clearly shows, the performance drops rapidly.
        At the original image scale, our LuNet model can almost match the mAP score of TriNet and already outperforms it when considering CMC scores.
        At an even smaller image size, LuNet significantly outperforms the pretrained model.
        Since the LuNet performance only drops by about $\sim 3\%$, the small images still hold enough data to perform ReID, but the rather rigid pretrained weights can no longer adjust to such a data change.
        This shows that pretrained models are not a solution for arbitrary tasks, especially when one wants to train lightweight models for small images.

\section{Discussion}\label{sec:discussion}
        We are not the first to use the triplet loss for person ReID.
        Ding~\etal~\cite{Ding15PR} and Wang~\etal~\cite{Wang16CVPR} use a batch generation and loss formulation which is very similar to our \textit{batch all} formulation.
        Wang~\etal~\cite{Wang16CVPR} further combine it with a pairwise verification loss.
        However, in the \textit{batch all} case, it was important for us to average only over the active triplets ($\lossnnz{BA}$), which they do not mention.
        This, in combination with their rather small networks, might explain their relatively low scores.
        Cheng~\etal~\cite{Cheng16CVPR} propose an ``improved triplet loss'' by introducing another pull term into the loss, penalizing large distances between positive images.
        This formulation is in fact very similar to the original one by Weinberger and Saul~\cite{Weinberger09JMLR}.
        We briefly experimented with a pull term, but the additional weighting hyper-parameter was not trivial to optimize and it did not improve our results.
        Several authors suggest learning attributes and ReID jointly~\cite{Khamis14ECCV,Su16ECCV,Lin17Arxiv,Schumann17CVPRW}, some of which integrate this into their embedding dimensions.
        This is an interesting research direction orthogonal to our work.
        Several other authors also defined losses over triplets of images~\cite{Paisitkriangkrai15CVPR,Shi16ECCV,Liu16MM}, however, they use losses different from the triplet loss we defend in this paper, possibly explaining their lower scores.
        Finally, FaceNet~\cite{Schroff15CVPR} uses a huge batch with moderate mining, which can only be done on the CPU,
        whereas we advocate hard mining in a small batch, which has a similar effect to moderate mining in a large batch,
        while fitting on a GPU and thus making training significantly more affordable.

\section{Conclusion}
    In this paper we have shown that, contrary to the prevailing belief, the triplet loss is an excellent tool for person re-identification.
    We propose a variant that no longer requires offline hard negative mining at almost no additional cost.
    Combined with a pretrained network, we set the new state-of-the-art on three of the major ReID datasets.
    Furthermore, we show that training networks from scratch can lead to very competitive scores.
    We hope that in future work the ReID community will build on top of our results and shift more towards end-to-end learning.

    \PAR{Acknowledgments.} This work was funded, in parts, by ERC Starting Grant project CV-SUPER (ERC-2012-StG-307432) and the EU project STRANDS (ICT-2011-600623).
    We would also like to thank the authors of the Market-1501 and MARS datasets.

{\small
\bibliographystyle{ieee}
\bibliography{abbrev_short,egbib}
}

\clearpage

\begin{appendix}

\noindent\begin{Large}
\hspace{-7.8pt}
\textbf{Supplementary Material}
\end{Large}

\section{Test-time Augmentation}
    As is good practice in the deep learning community~\cite{Krizhevsky12NIPS, He16CVPR}, we perform test-time augmentation.
    From each image, we deterministically create five crops of size $H \times W$: four corner crops and one center crop, as well as a horizontally flipped copy of each.
    The embeddings of all these ten images are then averaged, resulting in the final embedding for a person.
    Table~\ref{tab:augmentation} shows how five possible settings affect our scores on the Market-1501 dataset.
    As expected, the worst option is to scale the original images to fit the network input (first line), as this shows the network an image type it has never seen before.
    This option is directly followed by not using any test-time augmentation, \ie~just using the central crop.
    Simply flipping the center crop and averaging the two resulting embeddings already gives a big boost while only being twice as expensive.
    The four additional corner crops typically seem to be less effective, while more expensive, but using both augmentations together gives the best results.
    For the networks trained with a triplet loss, we gain about $3\%$ mAP, while the network trained for identification only gains about $2\%$ mAP.
    A possible explanation is that the feature space we learn with the triplet loss could be more suited to averaging than that of a classification network.

    \newcommand{\addaugmentdata}[1]{\csvreader[ column count=21, after line=, filter=\equal{\description}{#1}]
                                                {data/marsket_data.csv}
                                                {1=\description,  2=\singQueryMAP,
                                                                  4=\singQueryRankOne}
                                                {\np{\singQueryMAP}{}{}
                                                 && \np{\singQueryRankOne}{}{}
                                                }
                                        }

    \begin{table}[h]
       \footnotesize
       \setlength{\tabcolsep}{1pt}
       \setlength{\extrarowheight}{5pt}
       \renewcommand{\arraystretch}{0.75}
       \centering
       \begin{tabularx}{\linewidth}{p{1.6cm} Cp{2pt}C p{13pt} Cp{2pt}C p{13pt} Cp{2pt}C}
           \toprule[1pt]
            & \multicolumn{3}{c}{TriNet} && \multicolumn{3}{c}{LuNet} && \multicolumn{3}{c}{IDE (R) Ours}\\
           \cmidrule[0.5pt]{2-4} \cmidrule[0.5pt]{6-8} \cmidrule[0.5pt]{10-12}
            & mAP && rank-1 && mAP && rank-1 && mAP && rank-1\\
           \midrule[0.5pt]
           original       & \addaugmentdata{LaNet+no_resize} && \addaugmentdata{LuNet+no_resize} && \addaugmentdata{classification+max+no_resize+xqda}\\
           center         & \addaugmentdata{LaNet} && \addaugmentdata{LuNet} && \addaugmentdata{classification+max+xqda}\\
           center + flip  & \addaugmentdata{LaNet+flip} && \addaugmentdata{LuNet+flip} && \addaugmentdata{classification+max+flip+xqda}\\
           5 crops        & \addaugmentdata{LaNet+crop} && \addaugmentdata{LuNet+crop} && \addaugmentdata{classification+max+crop+xqda}\\
           5 crops + flip & \addaugmentdata{LaNet+flipcrop} && \addaugmentdata{LuNet+flipcrop} && \addaugmentdata{classification+max+flipcrop+xqda}\\
           \bottomrule[1pt]
       \end{tabularx}
       \caption{The effect of test-time augmentation on Market-1501.}
       \label{tab:augmentation}
    \end{table}

\section{Hard Positives, Hard Negatives and Outliers}
    \newcolumntype{Y}{>{\centering\arraybackslash}X}
    \newcommand{\imline}[6]{\includegraphics[width=1\linewidth]{pics/outliers/#1.jpg}&
                            \includegraphics[width=1\linewidth]{pics/outliers/#2.jpg}&
                            \includegraphics[width=1\linewidth]{pics/outliers/#3.jpg}&
                            \includegraphics[width=1\linewidth]{pics/outliers/#4.jpg}&
                            \includegraphics[width=1\linewidth]{pics/outliers/#5.jpg}&
                            \includegraphics[width=1\linewidth]{pics/outliers/#6.jpg}}

    \begin{figure}
        \scriptsize
        \setlength\tabcolsep{1pt}
        \begin{tabularx}{\linewidth}{@{}YYYYYY@{}}
            PID: 207 & PID: 207 & PID: 207 & PID: 209 & PID: 209 & PID: 209\\
            \imline{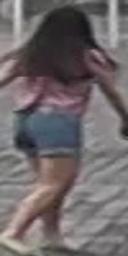}{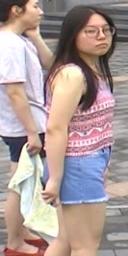}{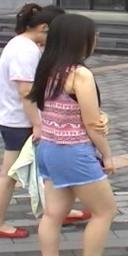}{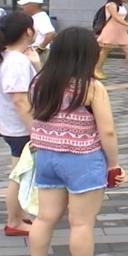}{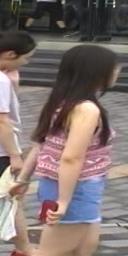}{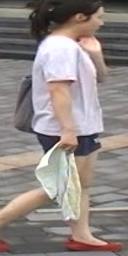}\\[5pt]
            PID: 1365 & PID: 1365 & PID: 1365 & PID: 1365 & PID: 1365 & PID: 1365\\
            \imline{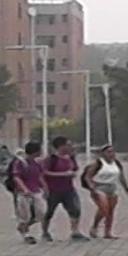}{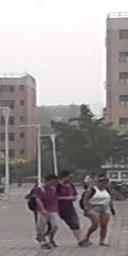}{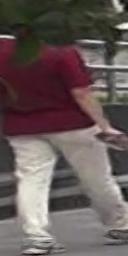}{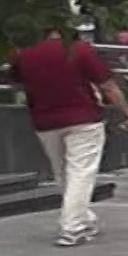}{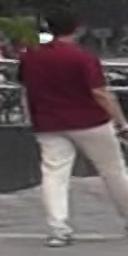}{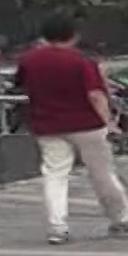}\\[5pt]
            PID: 516 & PID: 516 & PID: 516 & PID: 516 & PID: 516 & PID: 516\\
            \imline{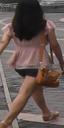}{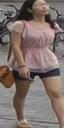}{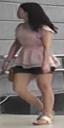}{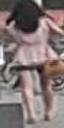}{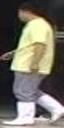}{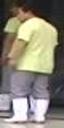}\\
        \end{tabularx}
        \caption{%
            Some outliers in the Mars (top two rows) and Market-1501 (bottom row) datasets.
            The first row shows high image overlap between tracklets of two persons.
            The second row shows a very hard example where a person was wrongly matched across tracklets.
            The last row shows a simple annotation mistake.
        }
        \label{fig:outliers}
    \end{figure}

    Figure~\ref{fig:outliers} shows several outliers in the Market-1501 and MARS datasets.
    Some issues in MARS are caused by the tracker-based annotations where bounding boxes sometimes span two persons and the tracker partially focuses on the wrong person.
    Additionally, some annotation mistakes can be found in both datasets; while some are obvious, some others are indeed very hard to spot!

    In Figure~\ref{fig:hard} we show some of the most difficult queries along with their top-3 retrieved images (containing hard negatives), as well as their two hardest positives.
    While some mistakes are easy to spot by a human, others are indeed not trivial, such as the first row in Figure~\ref{fig:hard}.

    \newcommand{\imhardfirst}[6]{\raisebox{-.5\height}{\includegraphics[width=1\linewidth,cfbox=#1 1pt 0pt]{pics/hard_x/#2.jpg}}&&&
                                 \raisebox{-.5\height}{\includegraphics[width=1\linewidth,cfbox=#3 1pt 0pt]{pics/hard_x/#4.jpg}}&
                                 \raisebox{-.5\height}{\includegraphics[width=1\linewidth,cfbox=#5 1pt 0pt]{pics/hard_x/#6.jpg}}&}

    \newcommand{\imhardsecond}[6]{\raisebox{-.5\height}{\includegraphics[width=1\linewidth,cfbox=#1 1pt 0pt]{pics/hard_x/#2.jpg}}& ~~$\dots$ &
                                  \raisebox{-.5\height}{\includegraphics[width=1\linewidth,cfbox=#3 1pt 0pt]{pics/hard_x/#4.jpg}}& ~~$\dots$ &
                                  \raisebox{-.5\height}{\includegraphics[width=1\linewidth,cfbox=#5 1pt 0pt]{pics/hard_x/#6.jpg}}}

    \begin{figure}
        \tiny
        \setlength\tabcolsep{1.5pt}
        \begin{tabularx}{\linewidth}{@{}Yc|cYYYcYcY@{}}
            PID: 1 &&& rank-1 & rank-2 & rank-3 && rank-1450 && rank-1967\\
            \imhardfirst{white}{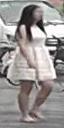}{better_green}{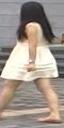}{better_red}{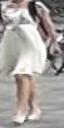}%
            \imhardsecond{better_green}{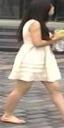}{better_green}{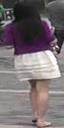}{better_green}{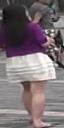}\\[35pt]%
            PID: 5 &&& rank-1 & rank-2 & rank-3 && rank-470 && rank-1130\\
            \imhardfirst{white}{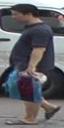}{better_red}{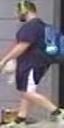}{better_red}{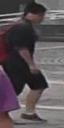}%
            \imhardsecond{better_red}{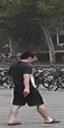}{better_green}{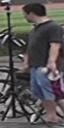}{better_green}{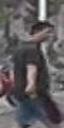}\\[35pt]%
            PID: 38 &&& rank-1 & rank-2 & rank-3 && rank-1510 && rank-2893\\
            \imhardfirst{white}{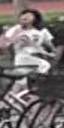}{better_green}{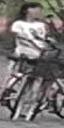}{better_red}{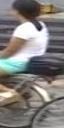}%
            \imhardsecond{better_red}{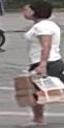}{better_green}{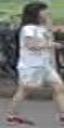}{better_green}{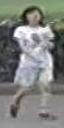}\\[35pt]%
            PID: 1026 &&& rank-1 & rank-2 & rank-3 && rank-2302 && rank-4611\\
            \imhardfirst{white}{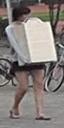}{better_green}{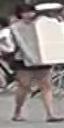}{better_green}{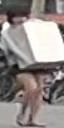}%
            \imhardsecond{better_red}{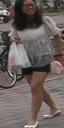}{better_green}{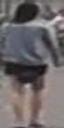}{better_green}{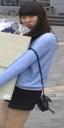}\\[35pt]%
            PID: 1151 &&& rank-1 & rank-2 & rank-3 && rank-1886 && rank-4127\\
            \imhardfirst{white}{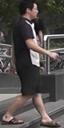}{better_green}{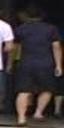}{better_red}{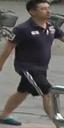}%
            \imhardsecond{better_red}{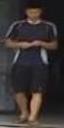}{better_green}{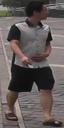}{better_green}{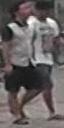}%
        \end{tabularx}
        \caption{%
            Some of the hardest queries.
            The leftmost column shows the query image, followed by the top 3 retrieved images and the two ground truth matches with the highest distance to the query images, \ie~the hardest positives.
            Correctly retrieved images have a green border, mistakes (\ie~hard negatives) have a red border.
        }
        \label{fig:hard}
    \end{figure}

\section{Experiments with Distractors}
    On top of the normal gallery set, the Market-1501 dataset provides an additional $500k$ distractors recorded at another time.
    In order to evaluate how such distractors affect performance, we randomly sample an increasing number of distractors and add them to the original gallery set.
    Here we compare to the results from Zheng~\etal~\cite{ZhengZ16Arxiv}.
    Both our models show a similar behavior to that of their ResNet-50 baseline.
    Surprisingly, our LuNet model starts out with a slightly better mAP score than the baseline and ends up just below it, while consistently being better when considering the rank-1 score.
    This might indeed suggest that the inductive bias from pretraining helps during generalization to large amounts of unseen data.
    Nevertheless, all models seem to suffer under the increasing gallery set in a similar manner, albeit none of them fails badly.
    Especially the fact that in $74.70\%$ of all single-image queries the first image out of $\num{519732}$ gallery images is correctly retrieved is an impressive result.

    \newcommand{\adddistplot}[4]{\addplot[mark=#3, color=#2] table [y=mAP #1, x expr={\thisrow{size}/1000}, col sep=comma] {data/market1501+500k.csv};
                \addlegendentry{#4}
                \addplot[mark=#3, color=#2, dashed] table [y=r1 #1, x expr={\thisrow{size}/1000}, col sep=comma,forget plot] {data/market1501+500k.csv};}

    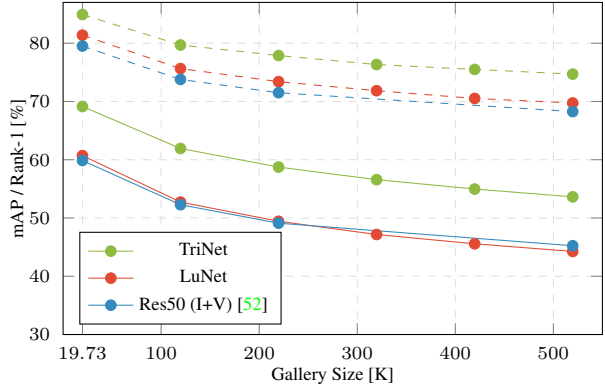
\begin{figure}[t]
        \begin{tikzpicture}
            \begin{axis}[
                ylabel style={font=\scriptsize,yshift=-18pt},
                xlabel style={font=\scriptsize,yshift=6pt},
                yticklabel style = {font=\scriptsize,xshift=-0.3ex},
                xticklabel style = {font=\scriptsize,yshift=-0.3ex},
                xtick={19.732,100,200,300,400,500},
                ytick={30,40,...,85},
                mark options={solid},
                grid=both,
                grid style={dashed,gray!25!white},
                xmin=0,
                xmax=550,
                ymin=30,
                ymax=87,
                width=1.05\linewidth,
                height=6cm,
                xlabel={Gallery Size [K]},
                ylabel={mAP / Rank-1 [\%]},
                legend style={legend pos=south west,font=\scriptsize}
            ]
                \adddistplot{LaNet}{better_green}{*}{TriNet}
                \adddistplot{LuNet}{better_red}{*}{LuNet}
                \adddistplot{LZ}{better_blue}{*}{Res50 (I+V)~\cite{ZhengZ16Arxiv}}
            \end{axis}
        \end{tikzpicture}
        \caption{500k distractor set results.
                 Solid lines represent the mAP score, dashed lines the rank-1 score.
                 See Supplementary Material for values.}
        \label{fig:500k_dist}
    \end{figure}

    For reproducibility of the 500k distractor plot (Fig.~\ref{fig:500k_dist}), Table~\ref{tab:distractors} lists the values of the plot.

    \newcommand{\adddistdata}[1]{\csvreader[ column count=21, after line=, filter=\equal{\size}{#1}]
                                                {data/market1501+500k.csv}
                                                {1=\size, 2=\MAPTrinet, 3=\RankOneTrinet
                                                        , 4=\MAPLunet, 5=\RankOneLunet
                                                        , 6=\MAPLZ, 7=\RankOneLZ}
                                                {\np{\MAPTrinet}{}{} && \np{\RankOneTrinet}{}{} &&
                                                 \np{\MAPLunet}{}{} && \np{\RankOneLunet}{}{} &&
                                                 \np{\MAPLZ}{}{} && \np{\RankOneLZ}{}{}
                                                }
                                        }

    \begin{table}[h]
       \footnotesize
       \setlength{\tabcolsep}{1pt}
       \setlength{\extrarowheight}{5pt}
       \renewcommand{\arraystretch}{0.75}
       \centering
       \begin{tabularx}{\linewidth}{p{1.6cm} Cp{2pt}C p{13pt} Cp{2pt}C p{13pt} Cp{2pt}C}
            \toprule[1pt]
             & \multicolumn{3}{c}{TriNet} && \multicolumn{3}{c}{LuNet} && \multicolumn{3}{c}{Res50 (I+V)~\cite{ZhengZ16Arxiv}}\\
            \cmidrule[0.5pt]{2-4} \cmidrule[0.5pt]{6-8} \cmidrule[0.5pt]{10-12}
            Gallery size & mAP && rank-1 && mAP && rank-1 && mAP && rank-1\\
            \midrule[0.5pt]
            ~~\num{19732} & \adddistdata{19732}\\
            \num{119732} & \adddistdata{119732}\\
            \num{219732} & \adddistdata{219732}\\
            \num{319732} & \adddistdata{319732}\\
            \num{419732} & \adddistdata{419732}\\
            \num{519732} & \adddistdata{519732}\\

           \bottomrule[1pt]
       \end{tabularx}
       \caption{Values for the 500k distractor plot.}
       \label{tab:distractors}
    \end{table}

\section{Notes on Network Training}
    Here we present and discuss several training-logs that display interesting behavior, as noted in the main paper.
    This serves as practical advice of what to monitor for practitioners who choose to use a triplet-based loss in their training.

    A typical training usually proceeds as follows: initially, all embeddings are pulled together towards their center of gravity.
    When they come close to each other, they will ``pass'' each other to join ``their'' clusters and, once this cross-over has happened,
    training mostly consists of pushing the clusters further apart and fine-tuning them.
    The \emph{collapsing} of training happens when the margin is too large and the initial spread is too small, such that the embeddings get stuck when trying to pass each other.

    Most importantly of all, if any type of hard-triplet mining is used, a stagnating loss curve by no means indicates stagnating progress.
    As the network learns to solve some hard cases, it will be presented with other hard cases and hence still keep a high loss.
    We recommend observing the fraction of active triplets in a batch, as well as the norms of the embeddings and all pairwise distances.

    \begin{figure}[t]
    \centering
        \begin{subfigure}{\linewidth}
            \includegraphics[width=1\linewidth]{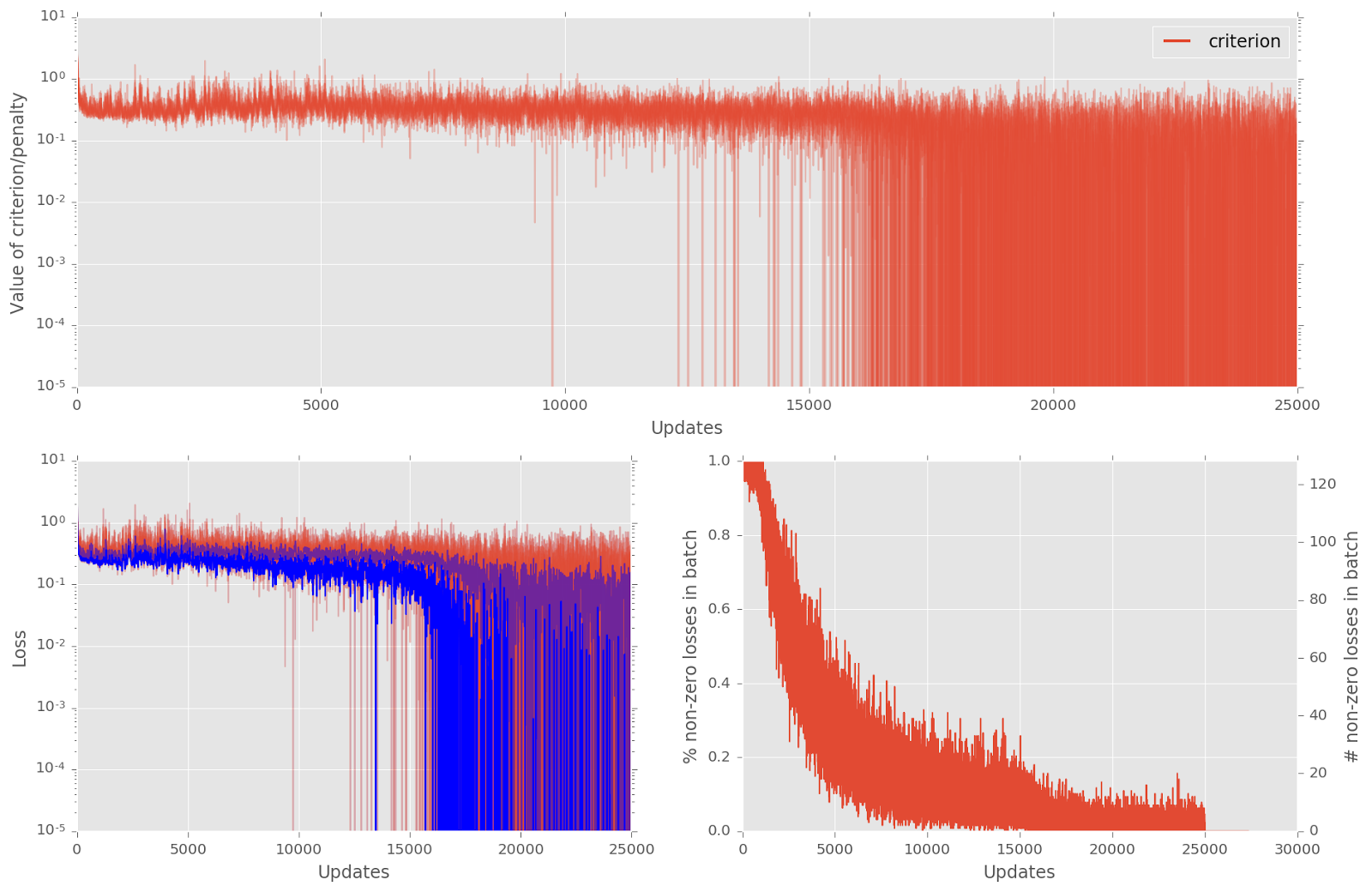}
            \caption{Training-log of the loss and active triplet count.}
        \end{subfigure}

        \begin{subfigure}{\linewidth}
            \includegraphics[width=1\linewidth]{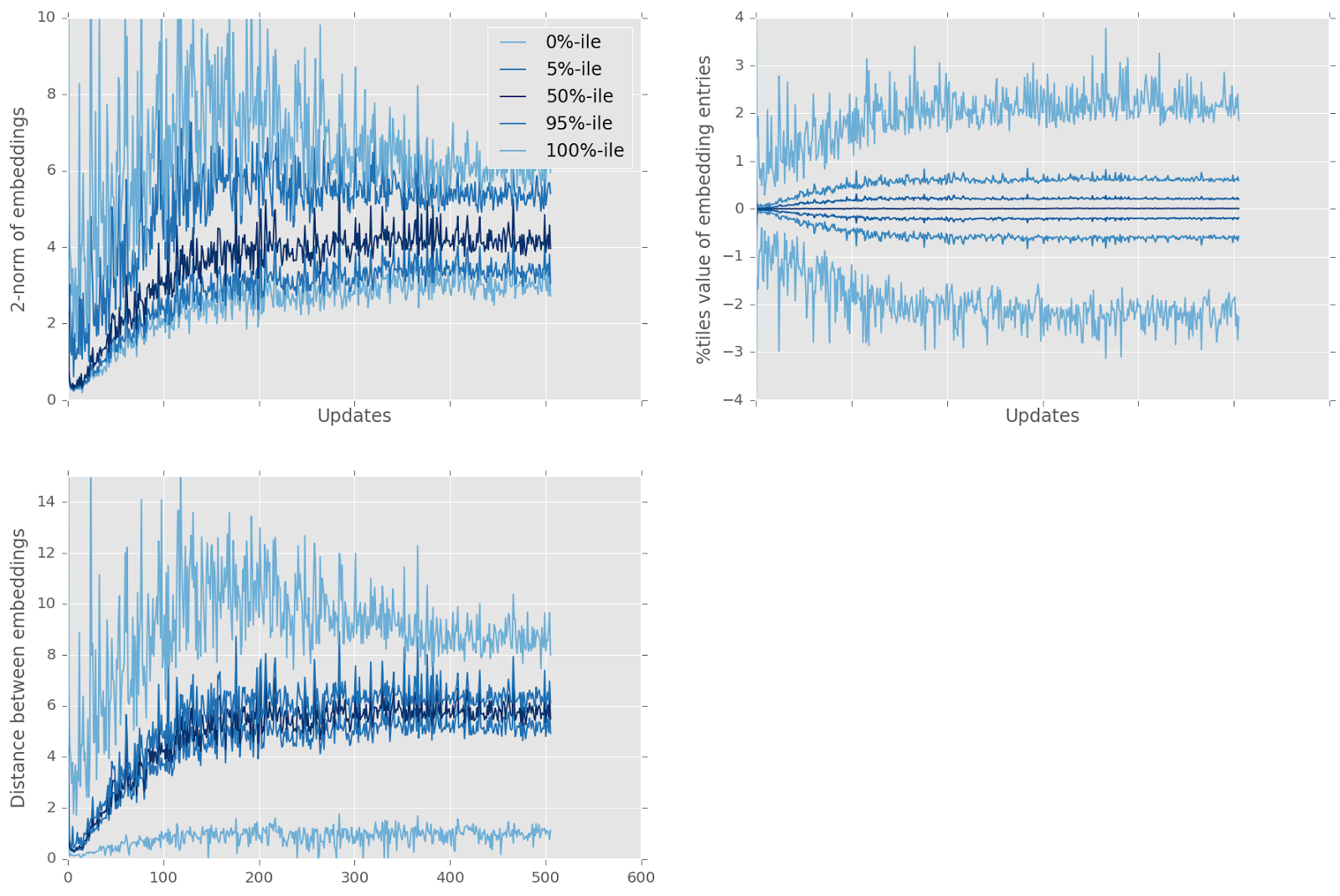}
            \caption{Training-log of the embeddings in the minibatch.}
        \end{subfigure}
        \caption{Training-log of LuNet on Market1501 using the \textit{batch hard} triplet loss with margin $0.2$. The embeddings stay bounded, as expected from a triplet formulation, and there is a lot of progress even when the loss stays seemingly flat.}
        \label{fig:lunet-market-m02}
    \end{figure}

    \begin{figure}[t]
    \centering
        \begin{subfigure}{\linewidth}
            \includegraphics[width=1\linewidth]{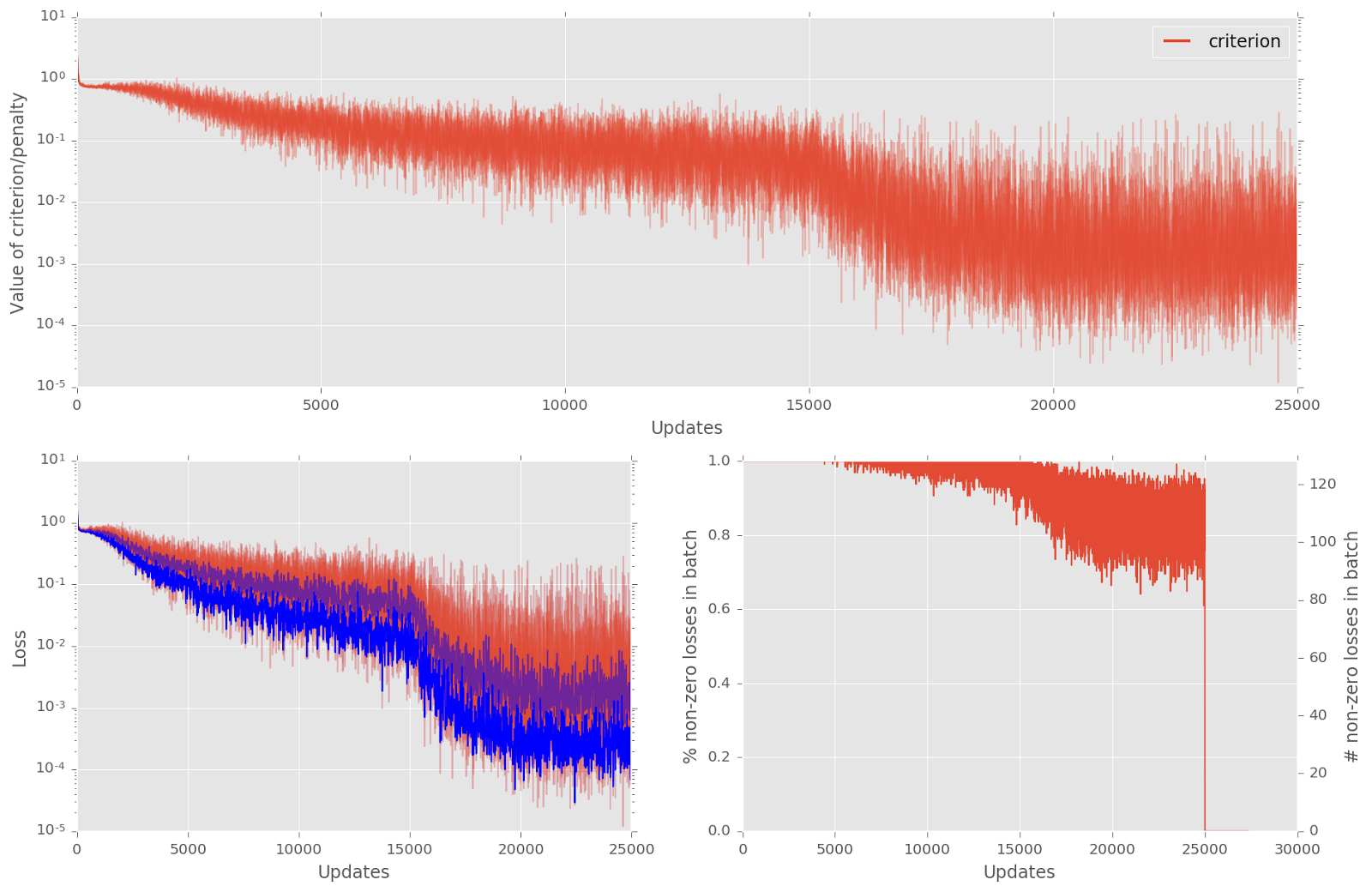}
            \caption{Training-log of the loss and active triplet count.}
        \end{subfigure}

        \begin{subfigure}{\linewidth}
            \includegraphics[width=1\linewidth]{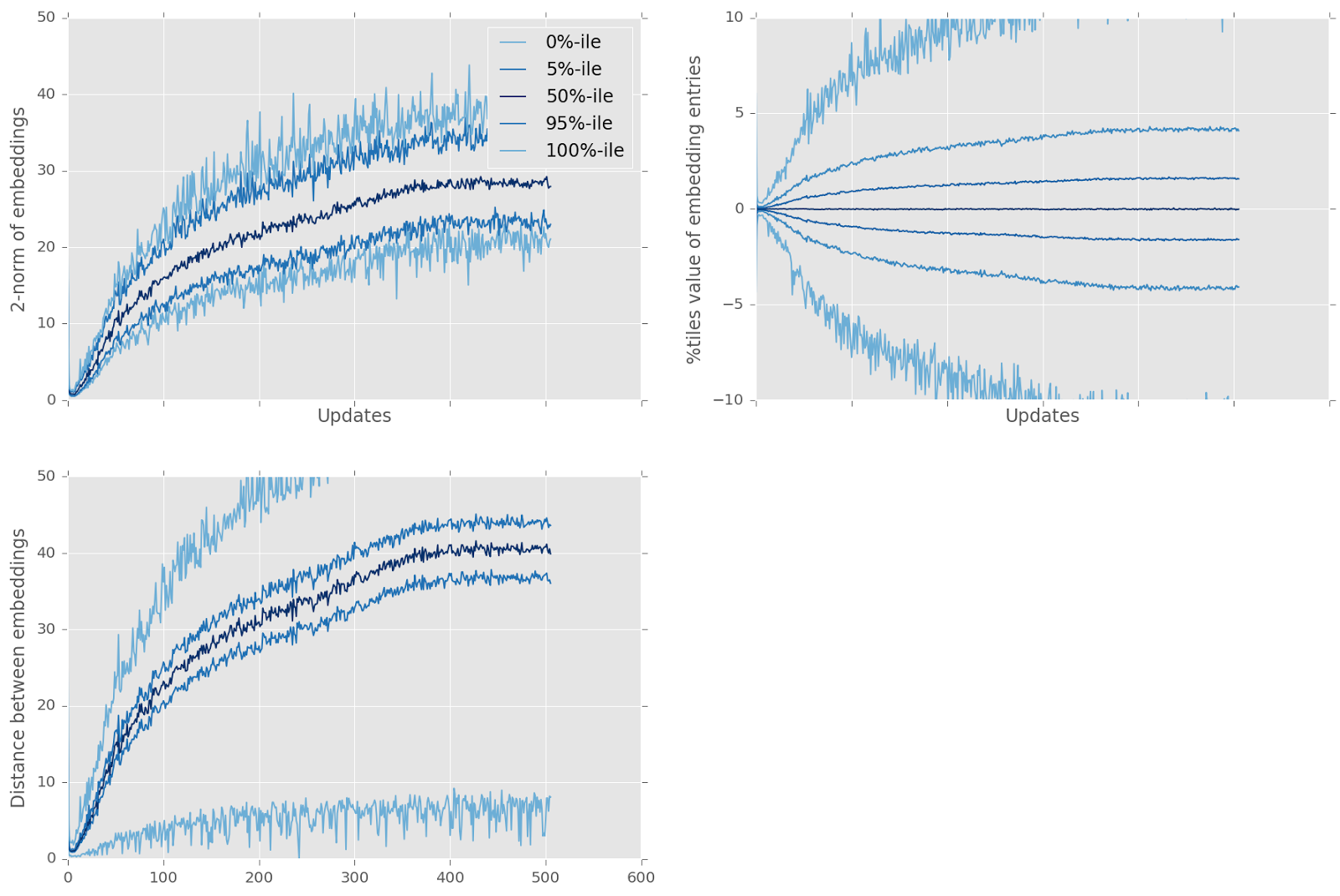}
            \caption{Training-log of the embeddings in the minibatch.}
        \end{subfigure}
        \caption{Training-log of LuNet on Market1501 using the \textit{batch hard} triplet loss with soft margin. The embeddings keep moving apart as even the loss shows a steady downward trend.\\
        }
        \label{fig:lunet-market-mNone}
    \end{figure}

\begin{figure}[t]
    \centering
        \begin{subfigure}{\linewidth}
            \includegraphics[width=1\linewidth]{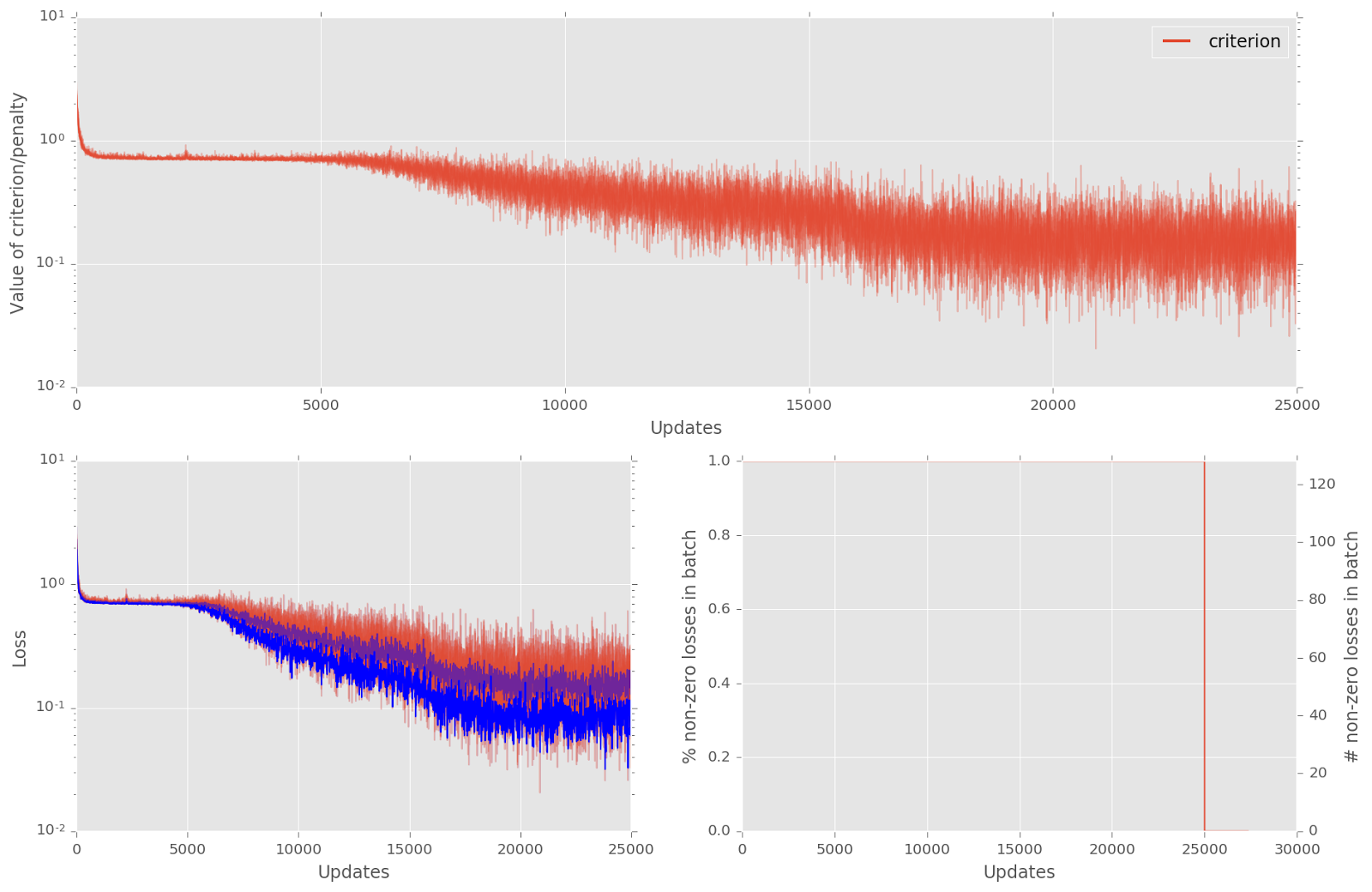}
            \caption{Training-log of the loss and active triplet count.}
        \end{subfigure}

        \begin{subfigure}{\linewidth}
            \includegraphics[width=1\linewidth]{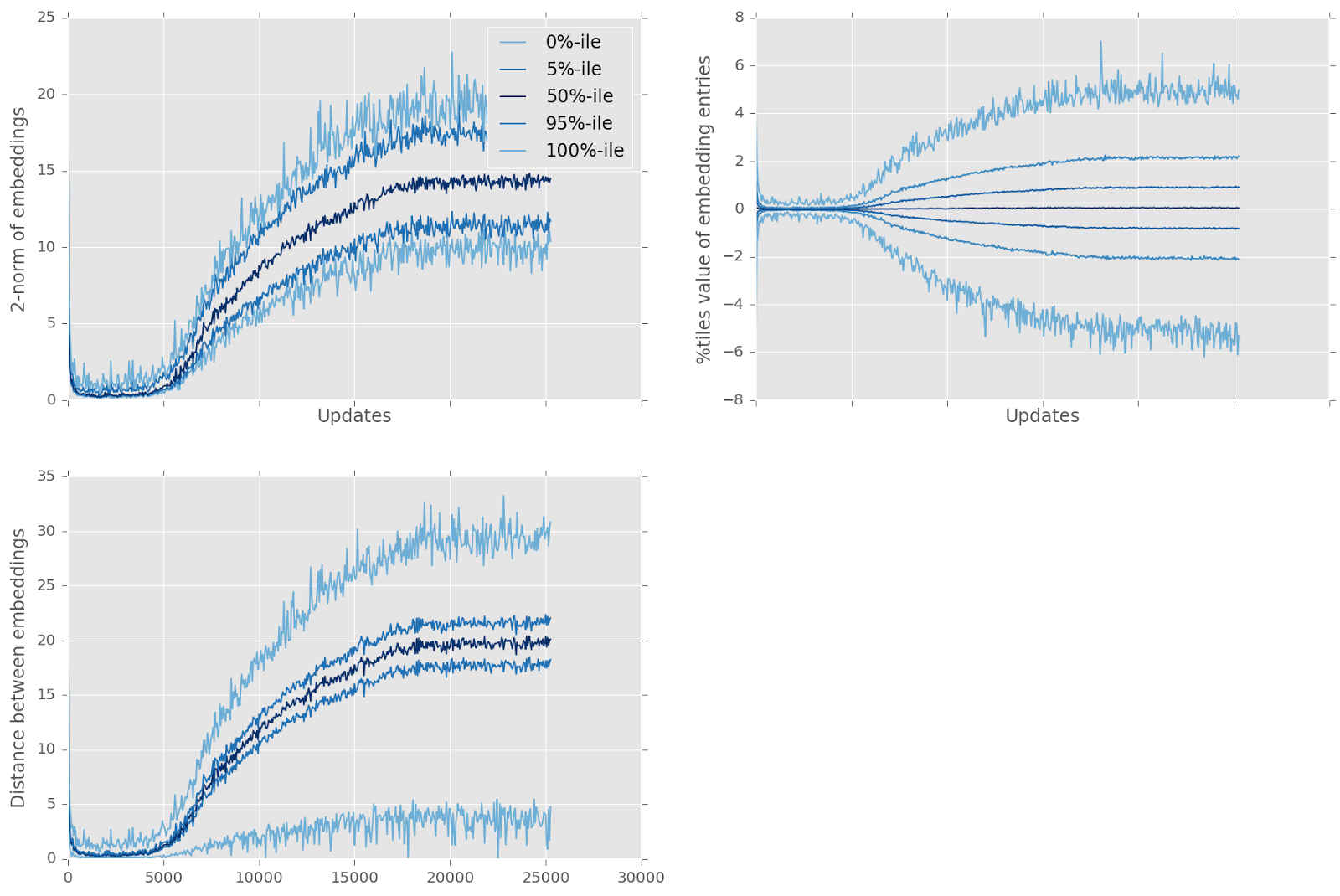}
            \caption{Training-log of the embeddings in the minibatch.}
        \end{subfigure}
        \caption{Training-log of a very small network on Market-1501 using the \textit{batch hard} triplet loss with soft margin. The difficult ``packed'' phase is clearly visible.}
        \label{fig:nanonet}
    \end{figure}

\begin{figure}[t]
    \centering
        \begin{subfigure}{\linewidth}
            \includegraphics[width=1\linewidth]{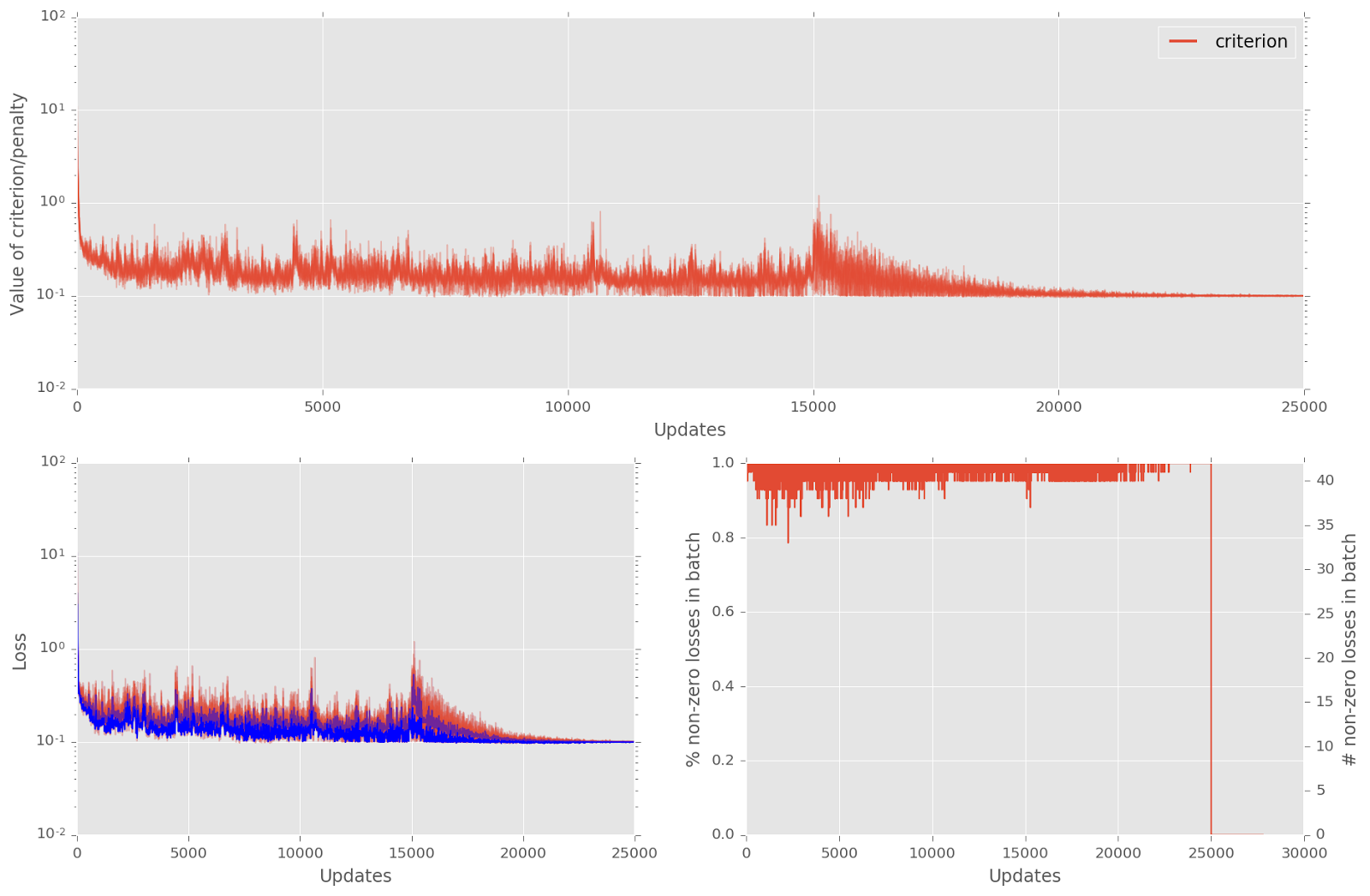}
            \caption{Training-log of the loss and active triplet count.}
        \end{subfigure}

        \begin{subfigure}{\linewidth}
            \includegraphics[width=1\linewidth]{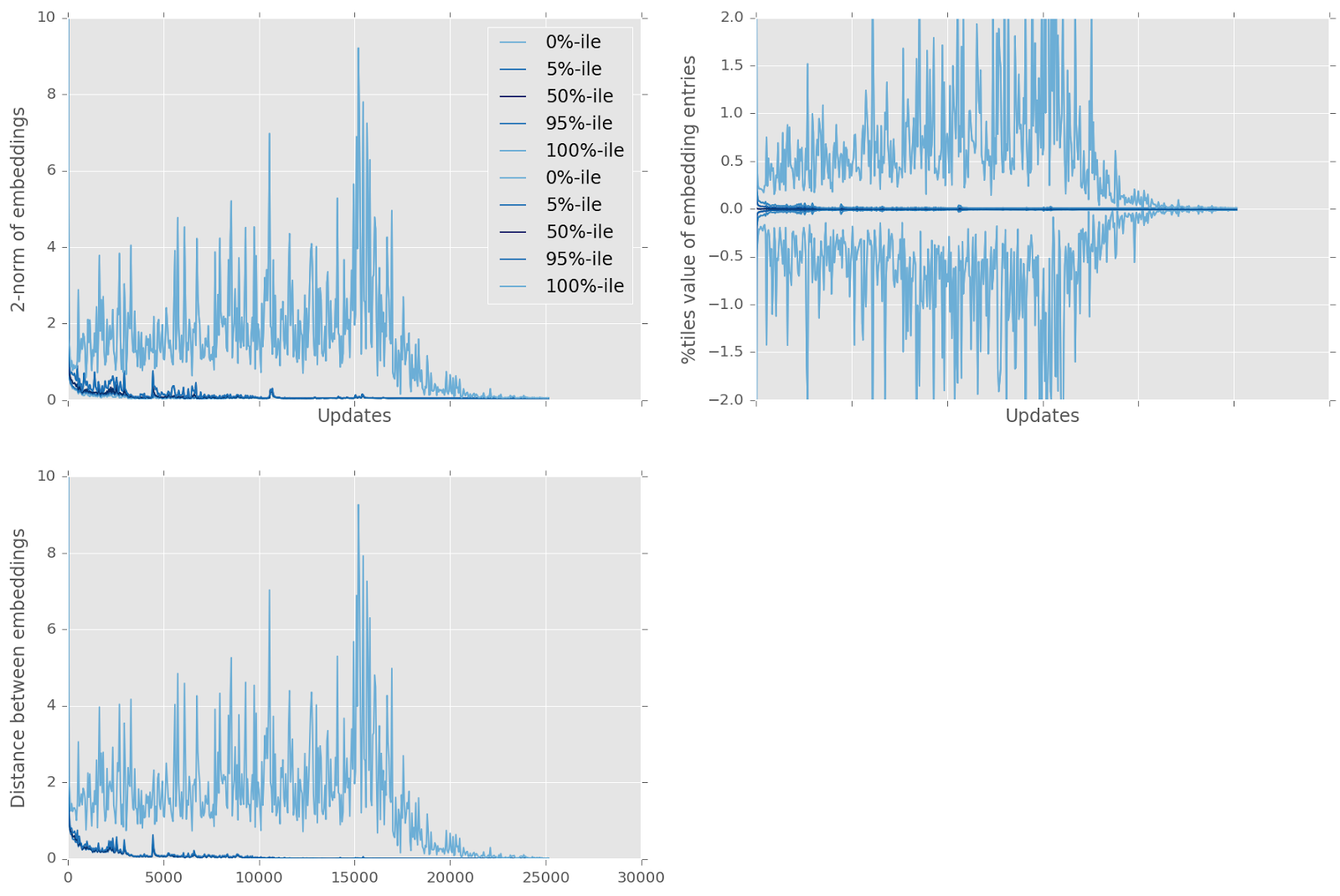}
            \caption{Training-log of the embeddings in the minibatch.}
        \end{subfigure}
        \caption{Training-log of LuNet on MARS when using offline hard mining (OHM) with margin $0.1$. This is one of the runs that collapsed and never got past the difficult phase.}
        \label{fig:lunet-ohm01}
    \end{figure}

    Figures~\ref{fig:lunet-market-m02},~\ref{fig:lunet-market-mNone},~\ref{fig:nanonet}, and~\ref{fig:lunet-ohm01} all show different training logs.
    Note that while they all share the x-axis since the number of updates was kept the same throughout the experiments, the y-axes vary for clearer visualization.
    First, the topmost plot in each Subfigure~(a) (``Value of criterion/penalty'') shows all per-triplet values of the optimization criterion (the triplet loss) encountered in each mini-batch.
    This is shown again in the plot below it on the left (``loss''), with an overlaid light-blue line representing the batch-mean criterion value, and an overlaid dark-blue line representing the batch's 5-percentile.
    To the right of it, the ``\% non-zero losses in batch'' shows how many entries in the mini-batch had non-zero loss; values are computed up to a precision of $10^{-5}$, which explains how it can be below $100\%$ in the soft-margin case.
    The single final vertical line present in some plots should be ignored as a plotting-artifact.

    Second, each Subfigure (b) (blue plots), monitors statistics about the embeddings computed during training.
    Different lines show $0$, $5$, $50$, $95$, and $100$-percentiles within a mini-batch, thus visualizing the distribution of values.
    The top-left plot, ``2-norm of embeddings'', shows the norms of the embeddings, thus visualizing whether the embedding-space shrinks towards 0 or expands.
    The top-right plot, ``\%tiles value of embedding entries'' shows these same statistics over the individual numeric entries in the embedding vectors.
    The only use we found for this plot is noticing when embeddings collapse to all-zeros vs.\ some other value.
    Finally, the bottom-left plot, ``Distance between embeddings'', is the most revealing, as it shows the same percentiles over all pairwise distances between the embeddings within a mini-batch.
    Due to a bug, the x-axis is unfortunately mislabeled in some cases.

    Let us now start by looking at the logs of two very successful runs: the LuNet training from scratch on Market-1501 with the \textit{batch hard} loss with margin $0.2$ and in the soft-margin formulation, see Fig.~\ref{fig:lunet-market-m02} and Fig.~\ref{fig:lunet-market-mNone}, respectively.
    The first observation is that, although they reach similar final scores, they learn significantly different embeddings.
    Looking at the embedding distances and norms, it is clear that the soft-margin formulation keeps pushing the embeddings apart, whereas the $0.2$-margin keeps the norm of embeddings and their distances bounded.
    The effect of exponentially decaying the learning-rate is also clearly visible: starting at $\num{15000}$ updates, both the loss as well as the number of non-zero entries in a mini-batch start to strongly decrease again, before finally converging from $\num{20000}$ to $\num{25000}$ updates.

    A network getting stuck only happened to us with too weak network architectures, or when using offline hard mining (OHM), the latter can be seen in Fig~\ref{fig:lunet-ohm01}.

    Next, let us turn to Fig.~\ref{fig:nanonet}, which shows the training-logs of a very small net (not further specified in this paper).
    We can clearly see that the network first pulls all embeddings towards their center of gravity, as evidenced by the quickly decreasing embedding norms, entries, as well as distances in the first few hundred updates.
    (More visible when zooming-in on a computer.)
    Once they are all close to each other, the networks really struggles to make them all ``pass each other'' to reach ``their'' clusters.
    As soon as this difficult phase is overcome, the embeddings are spread around the space to quickly decrease the loss.
    This is where the training becomes ``fragile'' and prone to collapsing: if the embeddings never pass each other, training gets stuck.
    This behavior can also be observed in Figures~\ref{fig:lunet-market-m02}~and~\ref{fig:lunet-market-mNone}, although to a much lesser extent, as the network is powerful enough to quickly overcome this difficult phase.


        \newcommand{\addfullmarskettableline}[3]{\csvreader[ column count=21, late after line=\\, filter=\equal{\description}{#1}]
                                                    {data/marsket_data.csv}
                                                    {1=\description,  2=\singQueryMAP,       3=\singQueryMAPBf,
                                                                      4=\singQueryRankOne,   5=\singQueryRankOneBf,
                                                                      6=\singQueryRankFive,  7=\singQueryRankFiveBf,
                                                                      8=\multQueryMAP,       9=\multQueryMAPBf,
                                                                     10=\multQueryRankOne,   11=\multQueryRankOneBf,
                                                                     12=\multQueryRankFive,  13=\multQueryRankFiveBf,
                                                                     14=\marsMAP,            15=\marsMAPBf,
                                                                     16=\marsRankOne,        17=\marsRankOneBf,
                                                                     18=\marsRankFive,       19=\marsRankFiveBf}
                                                    {#2 & #3
                                                     &  \np{\singQueryMAP}{\singQueryMAPBf}{}
                                                     && \np{\singQueryRankOne}{\singQueryRankOneBf}{}
                                                     && \np{\singQueryRankFive}{\singQueryRankFiveBf}{}
                                                     && \np{\multQueryMAP}{\multQueryMAPBf}{}
                                                     && \np{\multQueryRankOne}{\multQueryRankOneBf}{}
                                                     && \np{\multQueryRankFive}{\multQueryRankFiveBf}{}
                                                     && \np{\marsMAP}{\marsMAPBf}{}
                                                     && \np{\marsRankOne}{\marsRankOneBf}{}
                                                     && \np{\marsRankFive}{\marsRankFiveBf}{}
                                                    }
                                            }

        \begin{table*}[t]
            \setlength{\tabcolsep}{1pt}
            \setlength{\extrarowheight}{5pt}
            \renewcommand{\arraystretch}{0.75}
            \centering
            \begin{tabularx}{\textwidth}{p{4.2cm}cCp{1pt}Cp{1pt}Cp{8pt}Cp{1pt}Cp{1pt}Cp{18pt}Cp{1pt}Cp{1pt}C}
                \toprule[1pt]
                & \multirow{2}{25pt}{\hspace*{8pt}\rotatebox{90}{\hspace*{-10pt}Type}} & \multicolumn{5}{c}{Market-1501 SQ} && \multicolumn{5}{c}{Market-1501 MQ} && \multicolumn{5}{c}{MARS}\\
                \cmidrule[0.5pt]{3-7} \cmidrule[0.5pt]{9-13} \cmidrule[0.5pt]{15-19}
                 && mAP && rank-1 && rank-5 && mAP && rank-1 && rank-5 && mAP && rank-1 && rank-5\\
                \midrule[0.5pt]
                \addfullmarskettableline{LaNet+flipcrop}{TriNet}{E}
                \addfullmarskettableline{LuNet+flipcrop}{LuNet}{E}
                \addfullmarskettableline{classification+avg+flipcrop+xqda}{IDE (R) + ML ours }{I}
                \arrayrulecolor{lightgray}\midrule[0.25pt]\arrayrulecolor{black}
                \addfullmarskettableline{nullspace}{LOMO + Null Space~\cite{Zhang16CVPR}}{E}
                \addfullmarskettableline{gated siamese}{Gated siamese CNN~\cite{Varior16aECCV}}{V}
                \addfullmarskettableline{spindle net}{Spindle Net~\cite{Zhao17CVPR}}{I}
                \addfullmarskettableline{CAN}{CAN~\cite{Liu17TIP}}{E}
                \addfullmarskettableline{smooth manifold}{SSM~\cite{Bai17CVPR}}{-}
                \addfullmarskettableline{multi-loss}{JLML~\cite{Li17IJCAI}}{I}
                \addfullmarskettableline{svdnet}{SVDNet~\cite{Sun17ICCV}}{I}
                \addfullmarskettableline{GAN}{CNN + DCGAN~\cite{ZhengZ17ICCV}}{I}
                \addfullmarskettableline{multi scale}{DPFL~\cite{Chen17ICCVW}}{I}
                \addfullmarskettableline{discriminLZ}{ResNet 50 (I+V)$^\dagger$~\cite{ZhengZ16Arxiv}}{I+V}
                \addfullmarskettableline{mutual learning}{MobileNet+DML$^\dagger$~\cite{ZhangY17Arxiv}}{I}
                \addfullmarskettableline{DTL Trip}{DTL$^\dagger$~\cite{Geng16Arxiv}}{E}
                \addfullmarskettableline{DTL SID+PV}{DTL$^\dagger$~\cite{Geng16Arxiv}}{I+V}
                \addfullmarskettableline{attributes}{APR (R, 751)$^\dagger$~\cite{Lin17Arxiv}}{I}
                \addfullmarskettableline{alignment}{PAN$^\dagger$~\cite{ZhengZ17Arxiv}}{I}
                \addfullmarskettableline{mars}{IDE (C) + ML~\cite{ZhengL16ECCV}}{I}
                \addfullmarskettableline{latent parts}{Latent Parts (Fusion)~\cite{Li17CVPR}}{I}
                \addfullmarskettableline{reranking baseline}{IDE (R) + ML~\cite{Zhong17CVPR}}{I}
                \addfullmarskettableline{spatial temporal RNN}{Spatial-Temporal RNN~\cite{Zhou17CVPR}}{E}
                \addfullmarskettableline{compact appearance}{CNN + Video$^\dagger$~\cite{ZhangW17Arxiv}}{I}
                \arrayrulecolor{lightgray}\midrule[0.25pt]\arrayrulecolor{black}
                \addfullmarskettableline{LaNet+flipcrop+rerank}{TriNet (Re-ranked)}{E}
                \addfullmarskettableline{LuNet+flipcrop+rerank}{LuNet (Re-ranked)}{E}
                \addfullmarskettableline{classification+avg+flipcrop+xqda+rerank}{IDE (R) + ML ours (Re-ra.)}{I}
                \addfullmarskettableline{reranking}{IDE (R) + ML (Re-ra.)~\cite{Zhong17CVPR}}{I}
                \addfullmarskettableline{alignment reranking}{PAN (Re-ra.)$^\dagger$~\cite{ZhengZ17Arxiv}}{I}
                \bottomrule[1pt]
            \end{tabularx}
            \caption{%
                Scores on both the Market-1501 and MARS datasets.
                The top and middle contain our scores and those of the current state-of-the-art respectively.
                The bottom contains several methods with re-ranking~\cite{Zhong17CVPR}.
                The different types represent the optimization criteria, where I stands for identification, V for verification and E for embedding.
                All our scores include test-time augmentation.
                The best scores are bold.
                $^\dagger$: Concurrent work only published on arXiv.
            }
            \label{tab:marsket_results_full}
            \vspace*{-11pt}
        \end{table*}

\section{Extended Comparison Tables}
    We show extended versions of the two state-of-the-art comparison tables in the main paper.
    We add additional methods that were left out due to space reasons, or because the approaches are orthogonal to ours.
    The latter could be integrated with our approach in a straightforward manner.
    Market-1501 and Mars results are shown in Table~\ref{tab:marsket_results_full} and CUHK03 results in Table~\ref{tab:cuhk_results_full}.


\newpage
\mbox{}
\newpage
\mbox{}
\newpage

    \newcommand{\addfullcuhktableline}[3]{\csvreader[ column count=9, late after line=\\, filter=\equal{\description}{#1}]
                                                    {data/cuhk03.csv}
                                                    {1=\description, 2=\labeledRankOne,   3=\labeledRankOneBf,
                                                                     4=\labeledRankFive,  5=\labeledRankFiveBf,
                                                                     6=\detectedRankOne,  7=\detectedRankOneBf,
                                                                     8=\detectedRankFive, 9=\detectedRankFiveBf}
                                                    {#2 & #3
                                                     &  \np{\labeledRankOne}{\labeledRankOneBf}{}
                                                     && \np{\labeledRankFive}{\labeledRankFiveBf}{}
                                                     && \np{\detectedRankOne}{\detectedRankOneBf}{}
                                                     && \np{\detectedRankFive}{\detectedRankFiveBf}{}
                                                    }
                                            }

        \begin{table}[h!]
            \small
            \setlength{\tabcolsep}{1pt}
            \setlength{\extrarowheight}{5pt}
            \renewcommand{\arraystretch}{0.75}
            \centering
            \begin{tabularx}{\linewidth}{p{3.7cm}cCp{1pt}Cp{8pt}Cp{1pt}C}
                \toprule[1pt]
                & \multirow{2}{25pt}{\hspace*{8pt}\rotatebox{90}{\hspace*{-10pt}Type}} & \multicolumn{3}{c}{Labeled} && \multicolumn{3}{c}{Detected}\\
                \cmidrule[0.5pt]{3-5} \cmidrule[0.5pt]{7-9}
                 && r-1 && r-5 && r-1 && r-5\\
                \midrule[0.5pt]
                \addfullcuhktableline{LaNet+flipcrop}{TriNet}{E}
                \arrayrulecolor{lightgray}\midrule[0.25pt]\arrayrulecolor{black}
                \addfullcuhktableline{gated siamese}{Gated siamese CNN~\cite{Varior16aECCV}}{V}
                \addfullcuhktableline{domain guided dropout}{DGD*~\cite{Xiao16CVPR}}{I}
                \addfullcuhktableline{nullspace}{LOMO + Null Space~\cite{Zhang16CVPR}}{E}
                \addfullcuhktableline{smooth manifold}{SSM~\cite{Bai17CVPR}}{-}
                \addfullcuhktableline{CAN}{CAN~\cite{Liu17TIP}}{E}
                \addfullcuhktableline{latent parts}{Latent Parts (Fusion)~\cite{Li17CVPR}}{I}
                \addfullcuhktableline{spindle net}{Spindle Net*~\cite{Zhao17CVPR}}{I}
                \addfullcuhktableline{multi-loss}{JLML~\cite{Li17IJCAI}}{I}
                \addfullcuhktableline{svdnet}{SVDNet~\cite{Sun17ICCV}}{I}
                \addfullcuhktableline{GAN}{CNN + DCGAN~\cite{ZhengZ17ICCV}}{I}
                \addfullcuhktableline{multi scale}{DPFL~\cite{Chen17ICCVW}}{I}
                \addfullcuhktableline{DTL SID+PV}{DTL$^\dagger$~\cite{Geng16Arxiv}}{I+V}
                \addfullcuhktableline{discriminLZ}{ResNet 50 (I+V)$^\dagger$~\cite{ZhengZ16Arxiv}}{I+V}
                \bottomrule[1pt]
            \end{tabularx}
            \caption{%
                Scores on CUHK03 for TriNet and a set of recent top performing methods.
                The best scores are highlighted in bold.
                $^\dagger$: Concurrent work only published on arXiv.
                *: The method was trained on several additional datasets.
            }
            \label{tab:cuhk_results_full}
        \end{table}

\section{LuNet's Architecture}

\newcommand{\x}{\hspace{-2.5pt}\times\hspace{-2.5pt}}
\newcommand{\tf}[1]{\rule{0pt}{#1}}

    The details of the LuNet architecture for training from scratch can be seen in Table~\ref{table:architecture}.
    The input image has three channels and spatial dimensions $128\x64$.
    Most \emph{Res-block}s are of the ``bottleneck'' type~\cite{He16ECCV}, meaning for given numbers $n_1,n_2,n_3$ in the table, they consist of a $1\x1$ convolution from the number of input channels $n_1$ to the number of intermediate channels $n_2$, followed by a $3\x3$ convolution keeping the number of channel constant, and finally another $1\x1$ convolution going from $n_2$ channels to $n_3$ channels.
    Only the last Res-block, whose exact filter sizes are given in the table, is an exception to this.
    All ReLUs, including those in Res-blocks, are leaky~\cite{Maas13ICML} by a factor of $0.3$; although we do not have side-by-side experiments comparing the benefits, we expect them to be minor.
    All convolutional weights are initialized following He~\etal~\cite{He15ICCV}, whereas we initialized the final \emph{Linear} layers following Glorot~\etal~\cite{Glorot11AISTATS}.
    Batch-normalization~\cite{Ioffe15ICML} is essential to train such a network, and makes the exact initialization less important.

    \begin{table}[t]
        \begin{tabularx}{\linewidth}{p{50pt}X}
            \toprule[1pt]
            \tf{9pt}Type & Size \\

            \midrule[0.5pt]
            \arrayrulecolor{lightgray}
            \tf{9pt}Conv & $128\x7\x7\x3$ \\
            Res-block    & $128, 32, 128$ \\
            MaxPool      & pool $3\x3$, stride $(2\x2)$, padding $(1\x1)$ \\

            \midrule[0.5pt]
            \tf{9pt}Res-block & $128,32,128$ \\
            Res-block         & $128,32,128$ \\
            Res-block         & $128,64,256$ \\
            MaxPool           & pool $3\x3$, stride $(2\x2)$, padding $(1\x1)$ \\

            \midrule[0.5pt]
            \tf{9pt}Res-block & $256,64,256$ \\
            Res-block         & $256,64,256$ \\
            MaxPool           & pool $3\x3$, stride $(2\x2)$, padding $(1\x1)$ \\

            \cline{1-2}
            \tf{9pt}Res-block & $256,64,256$ \\
            Res-block         & $256,64,256$ \\
            Res-block         & $256,128,512$ \\
            MaxPool           & pool $3\x3$, stride $(2\x2)$, padding $(1\x1)$ \\

            \midrule[0.5pt]
            \tf{9pt}Res-block & $512,128,512$ \\
            Res-block         & $512,128,512$ \\
            MaxPool           & pool $3\x3$, stride $(2\x2)$, padding $(1\x1)$ \\

            \midrule[0.5pt]
            \tf{9pt}Res-block & $512\x(3\x3\x512), 128\x(3\x3\x512)$ \\
            Linear            & $1024\x512$ \\
            Batch-Norm        & $512$ \\
            ReLU              &  \\
            Linear            & $512\x128$ \\
            \arrayrulecolor{black}
            \bottomrule[1pt]
        \end{tabularx}
        \caption{The architecture of LuNet.}
        \label{table:architecture}
    \end{table}

    \begin{figure*}[t]
        \centering
        \includegraphics[width=\textwidth]{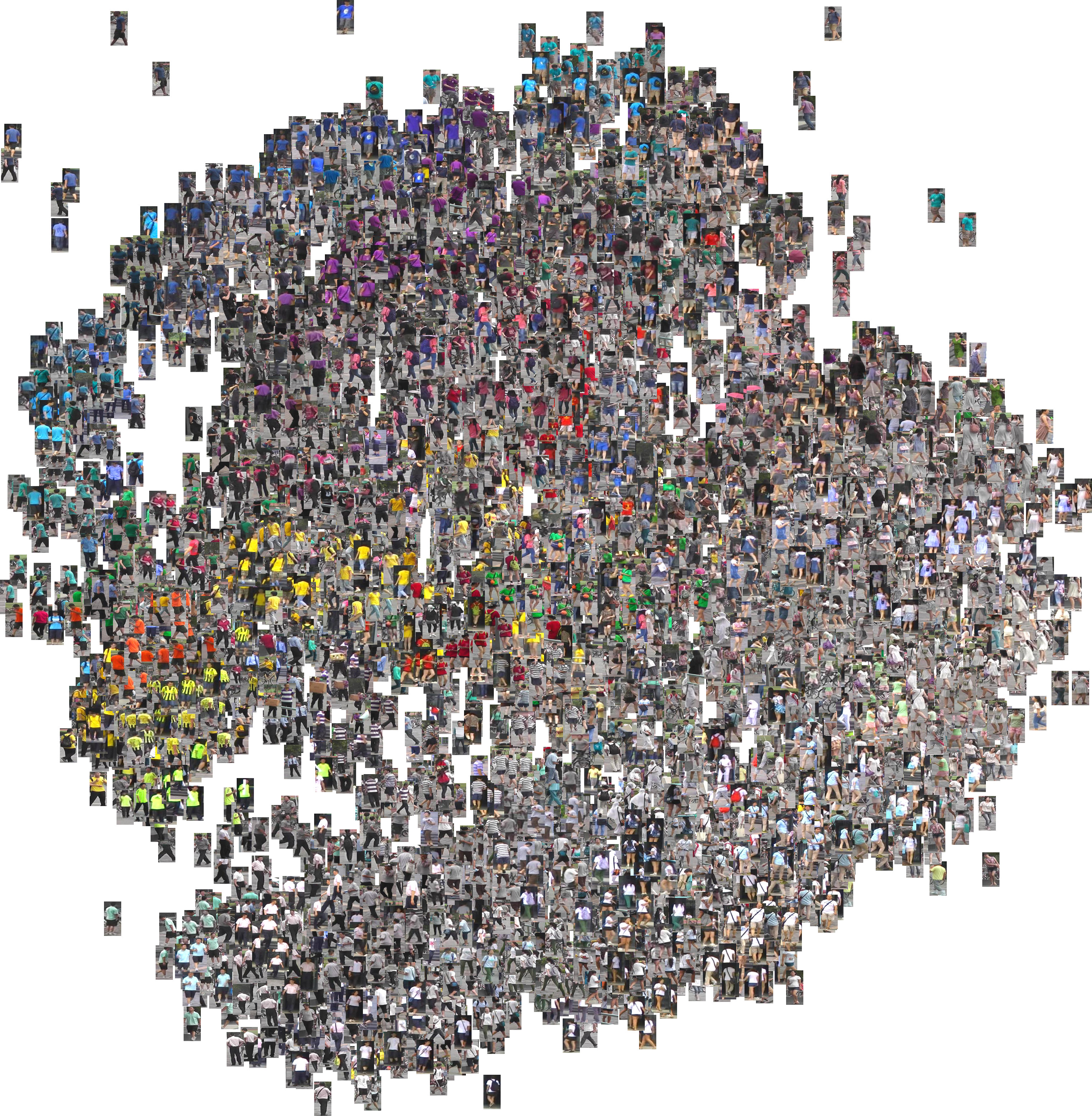}
        \caption{%
            Barnes-Hut t-SNE~\cite{Maaten14JMLR} of our learned embeddings for the Market-1501 test-set.
            Best viewed when zoomed-in.
        }
        \label{fig:tsne_full}
    \end{figure*}

\newpage
\mbox{}
\newpage

\section{Full t-SNE Visualization}
    Figure~\ref{fig:tsne_full} shows the full Barnes-Hut t-SNE visualization from which the teaser image (Fig.~1 in the paper) was cropped.
    We used a subset of \num{6000} images from the Market-1501 test-set and a perplexity of \num{5000} for this visualization.


\end{appendix}
\end{document}